\documentclass[sigconf]{acmart}

\usepackage{amsmath,amsfonts}
\usepackage{algorithmic}
\usepackage{algorithm}
\usepackage{array}

\usepackage{textcomp}
\usepackage{stfloats}
\usepackage{url}
\usepackage{verbatim}
\usepackage{graphicx}
\usepackage{booktabs}
\usepackage{multirow}
\usepackage{comment}
\usepackage{makecell}
\usepackage{subcaption}
\usepackage{hyperref}
\usepackage{cleveref}
\usepackage{url}

\AtBeginDocument{%
  \providecommand\BibTeX{{%
    \normalfont B\kern-0.5em{\scshape i\kern-0.25em b}\kern-0.8em\TeX}}}

\setcopyright{acmlicensed}
\copyrightyear{2024}
\acmYear{2024}
\acmDOI{10.1145/3664647.3681240}

\acmISBN{979-8-4007-0686-8/24/10}

\begin{document}

\title{Prompting Continual Person Search}


\author{Pengcheng Zhang}
\affiliation{%
  \institution{School of Computer Science and Engineering, State Key Laboratory of Complex \& Critical Software Environment, Jiangxi Research Institute, Beihang University}
  \city{Beijing}
  \country{China}}
\email{pengchengz@buaa.edu.cn}

\author{Xiaohan Yu}
\affiliation{%
  \institution{School of Computing, Macquarie University}
  \city{Sydney}
  \country{Australia}}
\email{xiaohan.yu@mq.edu.au}

\author{Xiao Bai}
\authornote{Corresponding author}
\affiliation{%
  \institution{School of Computer Science and Engineering, State Key Laboratory of Complex \& Critical Software Environment, Jiangxi Research Institute, Beihang University}
  \city{Beijing}
  \country{China}}
\email{baixiao@buaa.edu.cn}

\author{Jin Zheng}
\authornotemark[1]
\affiliation{%
  \institution{School of Computer Science and Engineering, State Key Laboratory of Complex \& Critical Software Environment, Jiangxi Research Institute, Beihang University}
  \city{Beijing}
  \country{China}}
\email{jinzheng@buaa.edu.cn}

\author{Xin Ning}
\affiliation{%
  \institution{Institute of Semiconductors, Chinese Academy of Sciences}
  \city{Beijing}
  \country{China}}
\email{ningxin@semi.ac.cn}

\begin{abstract}
  The development of person search techniques has been greatly promoted in recent years for its superior practicality and challenging goals. Despite their significant progress, existing person search models still lack the ability to continually learn from increasing real-world data and adaptively process input from different domains. To this end, this work introduces the continual person search task that sequentially learns on multiple domains and then performs person search on all seen domains. This requires balancing the stability and plasticity of the model to continually learn new knowledge without catastrophic forgetting. For this, we propose a \textbf{P}rompt-based C\textbf{o}ntinual \textbf{P}erson \textbf{S}earch (PoPS) model in this paper. First, we design a compositional person search transformer to construct an effective pre-trained transformer without exhaustive pre-training from scratch on large-scale person search data. This serves as the fundamental for prompt-based continual learning.
  On top of that, we design a domain incremental prompt pool with a diverse attribute matching module. For each domain, we independently learn a set of prompts to encode the domain-oriented knowledge. Meanwhile, we jointly learn a group of diverse attribute projections and prototype embeddings to capture discriminative domain attributes. By matching an input image with the learned attributes across domains, the learned prompts can be properly selected for model inference. Extensive experiments are conducted to validate the proposed method for continual person search.
  The source code is available at \url{https://github.com/PatrickZad/PoPS}.
\end{abstract}





\maketitle

\section{Introduction}

\begin{figure}[h]
  \centering
  \includegraphics[width=0.4\textwidth]{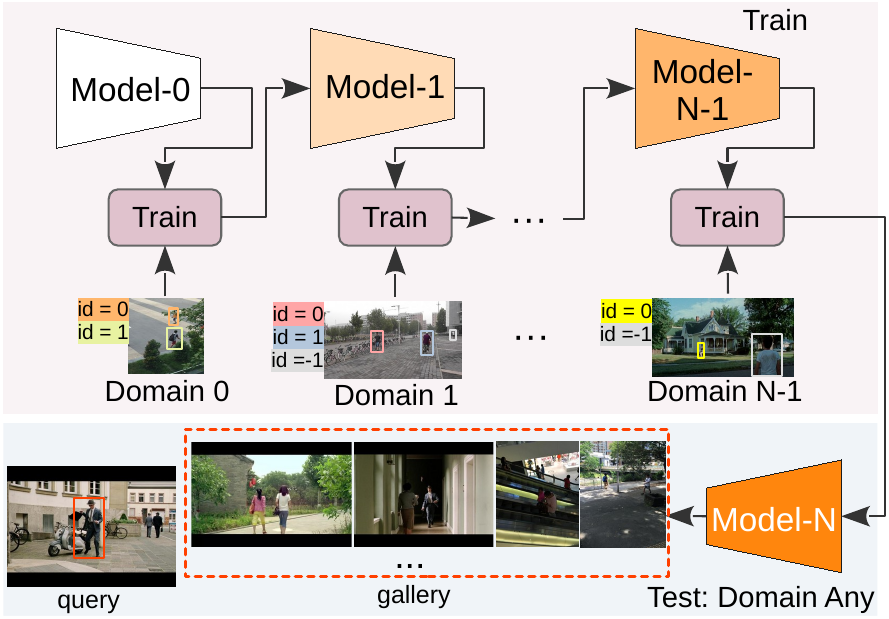}
  \caption{Illustration of the continual person feature problem.}
  \label{fig:def}
\end{figure}

Person search \cite{PS,prw,oim} aims to localize a target person in a gallery of uncropped scene images. It has attracted increasing research interest for its practicability and challenging goals. Existing works for person search have focused on boosting the performance under typical fully \cite{nae,seqnet,pstr,coat} or weakly \cite{cgps,rsiam,wang2023self} supervised scenarios, and exploring domain adaptation \cite{li2022domain} or generalization \cite{oh2024domain} methods. Despite their significant progress, these works learn only on a fixed and limited set of data while the real-world data is continually accumulating from different domains. To this end, we propose to explore the continual person search (CPS) problem that learns from sequentially incoming domains and adaptively completes the person search task for any learned domain (see Figure \ref{fig:def}).

A major challenge for enabling CPS is to balance the stability and plasticity of the model to consistently adapt to new domains without catastrophic forgetting of seen domains. Recent works \cite{l2p,dualprompt,coda,dualprompt,sprompts,gao2023unified} for continual image classification have drawn inspiration from visual prompt tuning \cite{vpt} to employ a frozen pre-trained transformer \cite{trans,vit} to guarantee model stability and expandable visual prompts to encode domain-oriented knowledge for plasticity. In this way, the inference of the models relies on properly selecting learned prompts to classify an image from any seen domain. As the transformers are pre-trained to learn image-level object visual representations, it is natural to incorporate those models to tackle the classification tasks. However, the models are not compatible with person search as the task requires jointly localizing and extracting instance-level features of persons in the scene image. A straightforward solution for this is to collect large-scale scene images of persons and pre-train a re-designed person search transformer from scratch. Yet this can be expensive due to (1) collecting and annotating sufficient data, \textit{e.g.} 14M images as in ImageNet-21K \cite{imagenet}, and (2) performing large-scale pre-training which may require a dozen large-memory GPUs running for several days.

Besides, previous prompt-based continual learning methods mainly tackle the class incremental learning \cite{l2p,dualprompt,coda,jung2023generating,gao2023unified} and domain incremental learning \cite{l2p,sprompts,jung2023generating} scenarios. Given that the former learns from sequential datasets with disjoint semantic space and the latter assumes all learning datasets share the same semantic space \cite{wang2024comprehensive,van2019three}, CPS is more closely related to the domain incremental learning scenario. However, the learning domains \cite{domainnet} in those works are with clear boundaries (\textit{e.g.} sketch image domain vs realistic image domain) which ease the adaptive selection of learned prompts during inference. In contrast, the domain gap between person search datasets can be ambiguous (\textit{e.g.} both CUHK-SYSU \cite{oim} and PRW \cite{prw} contain real-world images). This further raises a challenge to robustly capture the domain-specific attributes of an input image for properly selecting learned prompts to complete the person search task.

To tackle the aforementioned problems, we propose a \textbf{P}rompt-based C\textbf{o}ntinual \textbf{P}erson \textbf{S}earch (PoPS) model in this work. Specifically, we design a compositional person search transformer that employs an existing hierarchical vision transformer, \textit{e.g.} Swin \cite{swin}, and expand the transformer with a Simple Feature Pyramid \cite{simfpn} to enable person localization. The vision transformer is pre-trained on the ImageNet-22K \cite{imagenet} dataset and is publicly available. We then only train the Simple Feature Pyramid on a moderate number of person detection data to form a pre-trained person search transformer. This makes better use of existing pre-trained transformers to reduce the consumption of energy and resources of large-scale pre-training from scratch. It also requires less data to optimize such a lightweight detection sub-network.

On top of the proposed person search transformer, we design a domain incremental prompt pool with diverse attribute matching to enable CPS. The visual prompts are learned independently for each domain in the continual learning procedure similar to S-Prompts \cite{sprompts}. To capture domain-specific attributes for selecting learned prompts, we jointly learn a group of attribute projections and prototype embeddings. Similar to \cite{l2p,dualprompt,coda,sprompts}, a pre-trained transformer is employed to extract the global feature of an input image as the query embedding. We then use the attribute projection embedding to uncover the domain attribute in the query embedding and learn to match the attribute with the correlated attribute prototype embedding. By enforcing the attribute projection and prototype embeddings to be diverse, this diverse attribute matching mechanism is capable of capturing discriminative domain attributes. Therefore, the correlated prompts can be selected by measuring the similarity between a query embedding and learned attributes across different domains.

To summarize, this paper makes the following contributions:
\begin{itemize}
  \item We for the first time propose the continual person search problem. A Prompt-based Continual Person Search model is presented to consistently learn to adapt new person search tasks without catastrophic forgetting.
  \item  By constructing a compositional person search transformer, we reduce the cost of pre-training for prompt-based continual person search from scratch. A domain incremental prompt pool with diverse attribute matching is proposed to adaptively reuse learned prompts by measuring the similarity between input images and learned attributes across domains.
  \item Extensive experiments are conducted to understand the effectiveness of the proposed modules for continual person search.
\end{itemize}

\section{Related Work}

\textbf{Person Search.} The standard supervised learning of person search has been widely explored to achieve effective person search. \citeauthor{prw} \cite{prw} first explores combining popular person detector and person re-identification (ReID) models for person search, resulting in a two-step mechanism that first detects and crops person images and then retrieves a target across the cropped images. Following this mechanism, recent works obtained improved performance by enhanced person Re-ID features \cite{mgts,clsa} or designing target-conditioned person detectors \cite{igpn,tcts}. To improve the efficiency of the two-step paradigm, \citeauthor{oim} \cite{oim} proposed an end-to-end method to perform person search by a unified model.
Following works \cite{hoim,nae,dmrnet,galw} further explored to effectively balance the multiple training objectives as one-step person search is a multi-task learning problem. Other methods \cite{qeeps,ctxg,dcrnet} employed the context prior knowledge to match different persons across images. It is also practical to improve model efficiency by introducing lightweight detector architectures \cite{alignps,kdmot}, or boost the model performance by designing a stronger detection sub-network \cite{seqnet}. Inspired by recent advances in vision transformers \cite{trans,vit,ddetr}, recent works \cite{pstr,coat} obtained more discriminative person features with well-designed person search transformers. Recent works also explred the weakly supervised person search problem \cite{R-SiamNet,WCGPS,Wang_2023_ICCV} to train a person search model with only person bounding box annotations, and unsupervised domain adaptative person search \cite{uda_ps} that pre-trains on a labeled source domain and then adapts to an unlabeled target domain.


\textbf{Prompt-based Continual Learning.} To enable continual learning without a rehearsal buffer \cite{chaudhry2018efficient,hayes2019memory}, \citeauthor{l2p} \cite{l2p} proposed the first prompt-based continual learning method that designs a prompt pool with paired keys and prompts. The prompts help a pre-trained transformer to adapt to new tasks by visual prompt tuning \cite{vpt}, and the keys are learned to adaptively pick prompts for an input image. Based on this, DualPrompt \cite{dualprompt} jointly learned task-specific expert prompts and task-shared general prompts for prompt tuning. CODA-P \cite{coda} introduced a set of prompt components and implicitly learned the attention weight for fusing the prompt components, allowing adaptive weighted prompt summation instead of selection. LGCL \cite{lgcl} further introduced language guidance to learn a unified semantic embedding space for continual classification. DAP \cite{dap} instead learned a prompt generation module to construct a pool-free approach. Other works \cite{HiDe,tang2023prompt} analyzed the effect of pre-trained transformers and boosted the performance when using self-supervised pre-trained \cite{dino,mocov3} or moderate pre-trained transformers. Different from these works that mainly tackle class incremental learning \cite{wang2023self,van2019three}, S-Prompts \cite{sprompts} designed a simple yet effective method for domain incremental learning \cite{wang2024comprehensive,van2019three}.

\textbf{Lifelong ReID.} Another closely related topic to CPS is lifelong person re-identification (LReID) that learns from sequential ReID domains. For LReID, AKA\cite{pu2021aka} and MEGE \cite{pu2023jaka} designed effective knowledge graphs to adaptively accumulate and reuse learned knowledge. \citeauthor{sun2022patch} \cite{sun2022patch} proposed to adaptively choose patches for knowledge distillation. MRN \cite{pu2022meta} took a deep step into the proper batch normalization layers for LReID. Compared with these works, this work shares a similar spirit to encode and adaptively reuse learned knowledge by well-designed modules for multi-domain continual learning of person retrieval. This also eliminates the need for data-replay in many other LReID works \cite{yu2023lifelong,ge2022lifelong}. Meanwhile, this work also differs from those works as we simultaneously tackle the continual learning of person detection and present a unified framework to enable the continual learning of the two subtasks to collaboratively complete the CPS task.

\begin{figure*}[h]
  \centering
  \begin{subfigure}{0.32\textwidth}
    \includegraphics[width=\textwidth]{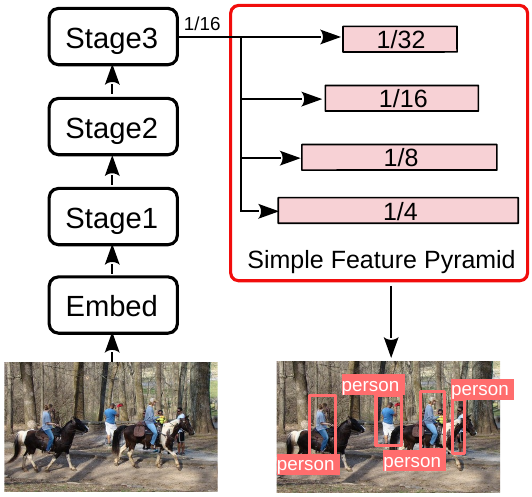}
    \caption{Pre-training by the detection task.}
    \label{fig:detp}
  \end{subfigure}
  \quad
  \quad
  \quad
  \quad
  \begin{subfigure}{0.32\textwidth}
    \centering
    \includegraphics[width=\textwidth]{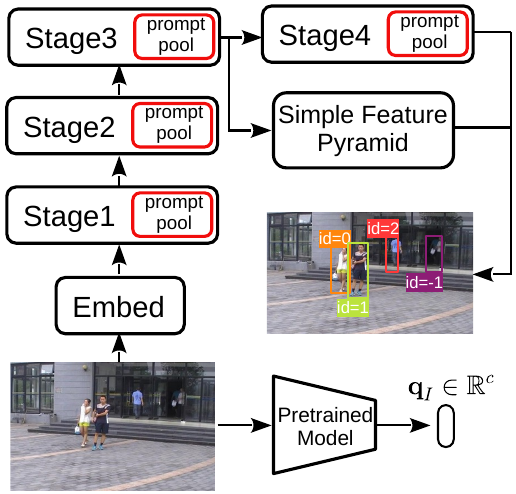}
    \caption{ The overall person search network.}
    \label{fig:promptc}
  \end{subfigure}
  \begin{subfigure}{0.9\textwidth}
    \centering
    \includegraphics[width=\textwidth]{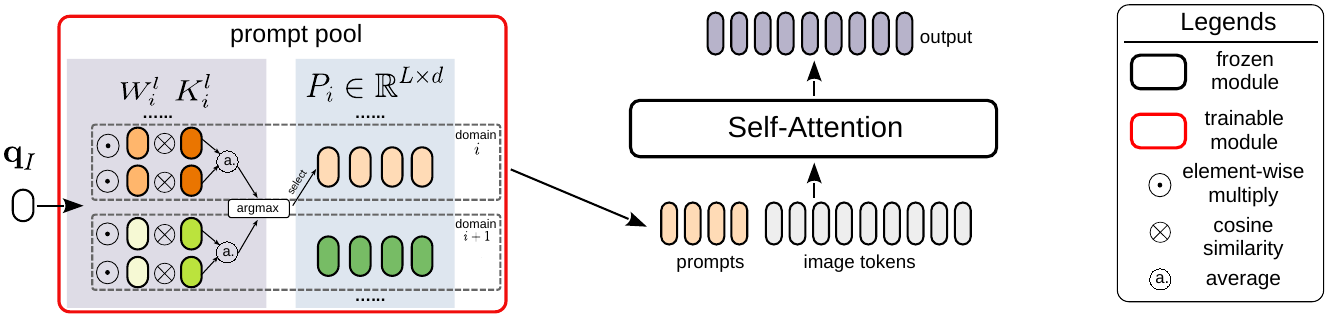}
    \caption{ The proposed prompt pool for each self-attention layer.}
    \label{fig:promptpool}
  \end{subfigure}
  \caption{Overview of the proposed method. (a) We first introduce a simple feature pyramid and perform detection pretraining to construct a pre-trained person search transformer. (b) We then inject the prompt pool for each transformer layer to enable CPS. The pre-trained model jointly extracts the global image feature $q_I$ which encodes domain-related information. (c) For each self-attention layer, the prompt pool independently learns the attribute projections $W_i^l$, attribute prototypes $K_i^l$, and prompts $P_i$ for each domain. $q_I$ is used to match with learned domain attributes and select the prompts for test. }
  \label{fig:cps}
\end{figure*}

\section{Method}
\subsection{Problem Definition and Overview} \label{overview}
In CPS, a unified model is trained sequentially on $T$ domains of person search and tested on an arbitrary domain without knowing the domain identity. Denoted by $\mathcal{D}=\{(D^i_{tr},D^i_{te})\}_{i=1}^T$ the learnable domains where $D^i_{tr}$ and $D^i_{te}$ are the train and test sets of domain $i$, respectively. $D^i_{tr}=\{\mathcal{X}^i_{tr},\mathcal{Y}^i_{tr}\}$ where $\mathcal{X}^i_{tr}$ indicates the training scene images and $\mathcal{Y}^i_{tr}$ contains identities and bounding boxes of persons in the images. Similarly, the test set is denoted by $D^i_{te}=\{\mathcal{X}^i_{te},\mathcal{Y}^i_{te}\}$. $\mathcal{Y}^i_{tr}\cap\mathcal{Y}^i_{te}=\emptyset$ and $D^i\cap D^{\neq i}=\emptyset$. During training, a unified model sequentially learns on $T$ train sets without accessing previously seen domains. During evaluation, the model is independently tested on the $T$ domains.

The overall architecture of the proposed PoPS model is depicted in Figure \ref{fig:cps}. We at first design a compositional person search transformer by expanding a hierarchical vision transformer \cite{swin} pre-trained on the widely-used ImageNet \cite{imagenet} data with a Simple Feature Pyramid \cite{simfpn} (see Figure \ref{fig:detp}). As the vision transformer is naturally capable of extracting person visual features, this enables the overall model to localize appeared persons in scene images and forms an effective person search network.
As the CPS problem is more closely related to the domain incremental learning task \cite{van2019three,wang2024comprehensive}, we propose to solve the problem by learning domain incremental visual prompts. Specifically, on top of the proposed person search network, we design a domain incremental prompt pool \cite{l2p,dualprompt,sprompts,coda} that independently learns visual prompts correlated with seen domains (see Figure \ref{fig:promptc}). By introducing diverse learnable attribute projection and prototype embeddings, the prompt pool learns to capture diverse domain attributes to select proper prompts at the test time (see Figure \ref{fig:promptpool}). On top of the designed modules, the overall continual learning procedure can be completed in two stages: (1) pre-training on a person detection task and (2) continual learning on incoming person search tasks.

In this section, we first describe how the compositional person search transformer process input scene images to complete the task in subsection \ref{model}, and then introduce the person detection pre-training and person search continual learning stages in subsections \ref{detp} and \ref{promptc}, respectively.

\subsection{Compositional Person Search Transformer} \label{model}

Largely pre-trained vision transformers \cite{trans,vit} are vital for prompt-based continual learning \cite{l2p,dualprompt,coda,sprompts}. Yet the models are not directly applicable to the person search task for the lack of person localization modules. It is also time- and resource-consuming to reformulate and pre-train a person search transformer from scratch. To this end, we employ a typical hierarchical vision transformer, \textit{i.e.} Swin \cite{swin}, pre-trained on the ImageNet \cite{imagenet} data and expand the transformer with a Simple Feature Pyramid \cite{simfpn} to enable person localization.

As in Figure \ref{fig:detp}, an input image $\mathbf{I}\in \mathbb{R}^{H_0\times W_0\times 3}$ is partitioned into multiple equally-sized patches and each patch is projected into a high-dimensional vector by the `Patch Embed' layer, resulting in an intermediate image feature map $\mathbf{F}_{img}\in \mathbb{R}^{H\times W\times C}$. Afterward, Swin introduces 4 stages where each stage performs downsampling and consecutive window-based multi-head self-attention \cite{trans} on their inputs to produce deep visual representations. On top of that, we introduce the Simple Feature Pyramid \cite{simfpn} to enable the overall model for person localization. Following \cite{simfpn}, we build the Simple Feature Pyramid on the 16 times downsampled feature map $\mathbf{F}'_{img}\in \mathbb{R}^{H'\times W'\times C'}$, \textit{i.e.} the output of `Stage3'. By applying convolutions of strides $\left\{2,1,\frac{1}{2},\frac{1}{4}\right\}$ on $\mathbf{F}'_{img}$ in parallel, where the fractional strides indicate deconvolutions, we obtain image feature maps of scales $\left\{\frac{1}{32},\frac{1}{16},\frac{1}{8},\frac{1}{4}\right\}$. Based on the multi-scale features maps, we follow Mask R-CNN \cite{maskrcnn} to predict person bounding boxes and detection confidences.

Different from the image classification \cite{swin,mvit} or person re-identification tasks \cite{agw,hat} that directly extract object features from object-centered images, the person search task requires to extract features of instances in the scene images. To this end, we dissect Swin into two parts, \textit{i.e.} `Patch Embed' to `Stage3' blocks that process the integral image feature maps, and the `Stage4' block that refines instance-wise person feature maps. 
According to the person bounding boxes predicted by the detection sub-network, we use ROIAlign \cite{maskrcnn} to extract interpolated feature maps from the output of `Stage3' within the boxes as person feature maps. We then employ the `Stage4' block on the extracted feature maps to obtain refined person feature maps $\mathbf{f}_{roi}\in\mathbb{R}^{h\times\ w\times d}$. Afterward, we follow the practice in Swin to conduct global average pooling on $\mathbf{f}_{roi}$ and use the pre-trained Layer Normalization layer to produce final 1D person features $\mathbf{v} \in \mathbb{R}^d$.


\subsection{Pre-training by Person Detection} \label{detp}

Although the compositional person search transformer makes a pre-trained transformer applicable for person search, the newly added modules are randomly initialized and thus contain no prior knowledge for subsequent continual learning. To tackle this problem, we propose to pre-train only the detection sub-network to construct a pre-trained person search transformer. As the pre-trained Swin \cite{swin} is frozen in this stage, the pre-trained visual feature space is left unchanged for person feature extraction and the detection sub-network only needs to predict person locations from the pre-trained image features. It is also worth noting that the number of the learnable parameters in this mechanism is relatively small, thus the pre-training can be completed with less data.

Specifically, we combine the training set of CrowdHuman \cite{crowdhuman} and images containing humans in the training set of MSCOCO \cite{coco} to form a person detection dataset similar to \citeauthor{shuai2022id} \cite{shuai2022id}. In total, this collection presents 79,115 scene images that contain nearly 0.6M person instances for training the model. Compared with the pre-training (using ImageNet-21K \cite{imagenet} that contains 14M images) of the widely used vision transformers in prompt-based continual learning methods \cite{l2p,dualprompt,coda,sprompts}, the proposed pre-training by person detection is more data-efficient.

Based on the collected data, we pre-train the model by conducting person detection to form a pre-trained person search transformer. As is mentioned in Subsection \ref{model}, we employ the `Patch Embed' to `Stage3' blocks of the pre-trained Swin \cite{swin} to extract image feature maps and send the feature maps to the detection sub-network to detect appeared persons. The overall training objective is formulated as
\begin{equation}
  \mathcal{L}_{det}=\mathcal{L}_{reg}+\mathcal{L}_{cls}+\mathcal{L}_{reg}^{rpn}+\mathcal{L}_{cls}^{rpn}
  \label{det_loss}
\end{equation}
where $\mathcal{L}_{reg}$ and $\mathcal{L}_{cls}$ are the bounding box regression and classification losses of the detection head \cite{maskrcnn}, $\mathcal{L}_{reg}^{rpn}$ and $\mathcal{L}_{cls}^{rpn}$ are those of the Region Proposal Network \cite{fasterrcnn}.

We also note that the data used for this pre-training stage shares similar spirits with that in \cite{shuai2022id}. However, \citeauthor{shuai2022id} \cite{shuai2022id} employs the data to pre-train a convolutional person search network by a self-supervised person similarity learning framework. The pre-trained model can be fine-tuned for a specific downstream task to achieve improved person search performance. Different from \citeauthor{shuai2022id} \cite{shuai2022id}, we focus on efficiently pre-training a person search transformer by a person detection task with the collected data. The pre-trained model is used to enable prompt-based continual learning without fine-tuning.

\subsection{Continual Learning for Person Search} \label{promptc}
A key target in continual learning is to balance the stability and plasticity of the model for accumulating knowledge of new tasks without forgetting that of learned tasks. For this, recent advanced continual learning methods \cite{sprompts,jung2023generating,coda,l2p,dualprompt} have explored a prompt-based mechanism that exploits a largely pre-trained vision transformer with an incrementally learned prompt \cite{vpt} pool. In this way, the pre-trained transformer is fixed and a set of learnable prompts are injected into the input sequence to the self-attention layer \cite{trans,vit} to adapt to a new task. Thus the continual learning problem can be solved by continually learning new visual prompts for new tasks during training and adaptively reusing proper prompts during inference. Inspired by this mechanism, we design a domain incremental prompt pool with diverse attribute matching to enable CPS.

\textbf{Domain incremental prompt pool.} On top of the pre-trained compositional person search transformer, we independently learn domain-oriented visual prompts during continual learning similar to S-Prompts \cite{sprompts}. Specifically for a domain $i$, we set a sequence of learnable prompts $P_{i}^l\in \mathbb{R}^{L\times d}$, where $L$ is the number of prompts, for the $l$-th self-attention layer as in Figure \ref{fig:promptpool}. Different from previous works \cite{l2p,dualprompt,coda,sprompts} that employ ViT \cite{vit} with global self-attention layers, Swin \cite{swin} partitions the input feature map into local windows and performs window-based self-attention \cite{swin} to reduce the computation complexity. Thus for the $l$-th Swin transformer layer, we duplicate the correlated prompts $P_{i}^l$ for each partitioned window and conduct self-attention as
\begin{equation}
  \hat{\mathbf{z}}^l=\mathrm{MHSA}(\mathrm{CAT}(P_i^l,\mathbf{z}^{l-1}))[L:]
\end{equation}
where $\mathrm{MHSA}$ refers to multi-head self-attention and $\mathrm{CAT}$ is the concatenation operation. $\mathbf{z}^{l-1}$ is the flattened image feature map within a local window. As $\mathrm{MHSA}$ keeps the length of the input sequence, we use $[L:]$ to preserve only the output corresponding to the input image feature tokens, leaving the spatial shape of the overall image feature maps unchanged for subsequent blocks. By applying the learnable prompts, we jointly train the model in Figure \ref{fig:promptc} with the detection loss in Equation \ref{det_loss} and the widely used Online-Instance-Matching (OIM) loss \cite{oim} $\mathcal{L}_{oim}$ for each incoming person search domain.

\textbf{Diverse attribute matching.} As the domain identity of a test task is unavailable during inference, the learned visual prompts should be adaptively selected for inference. To tackle this problem, we design a diverse attribute matching module to measure the similarity between an input image and seen domains. Following previous prompt-based continual learning methods \cite{l2p,dualprompt,coda, sprompts}, we duplicate the pre-trained Swin \cite{swin} to extract the global visual feature as $\textbf{q}_i=F(I_i), \textbf{q}_i\in\mathbb{R}^c$ of an input image $I_i$ belonging to domain $i$. As the pre-trained model is agnostic to the incoming person search tasks, the visual feature $\textbf{q}_{i}$ implicitly encodes unbiased attributes of domain $i$. To capture discriminative domain features, \textit{i.e.} domain attributes, we bind a group of $N$ learnable embeddings $K_i^l=\left\{\mathbf{k}_i^j\in\mathbb{R}^c|j=1,2,\dots N\right\}$ with the visual prompts $P_i^l$ as attribute prototypes of domain $i$. By maximizing the similarity between $\textbf{q}_{i}$ and $\mathbf{k}_i^j$ similar to L2P \cite{l2p}, the attribute embeddings are optimized to match the attributes of domain $i$ during training and the learned prompts can be selected by matching the input image with learned domain attributes during inference. However, this may cause redundancy between different attribute embeddings as the multiple learnable prototypes can easily overfit the same prominent domain attribute. As the domain scenarios can be highly similar between different person search datasets, it also requires learning more diverse domain attributes to mostly uncover the differences between different domains.

For this, we further attach a group of learnable attribute projection embeddings $W_i^l=\left\{\mathbf{w}_i^j\in\mathbb{R}^c|j=1,2,\dots N\right\}$ to $K_i^l$. $W_i^l$ can be regarded as learned channel attention \cite{senet} to emphasize a certain attribute of domain $i$ in an input image. For training on domain $i$, given an input scene image $I_i$, we calculate the similarity between the image global feature $\textbf{q}_{i}=F(I_i)$ and the attribute prototype $\mathbf{k}_i^j$ as
\begin{equation}
  a_i^j=\textbf{q}_{i}\odot\mathbf{w}_i^j\otimes\mathbf{k}_i^j
\end{equation}
where $\odot$ denotes channel-wise multiplication and $\otimes$ stands for calculating cosine similarity. The domain attribute learning loss is thus formulated as
\begin{equation}
  \mathcal{L}_{attr}=\sum_{j=1}^{N}\left(1-a_i^j\right)
\end{equation}
Besides, both $\left\{\mathbf{w}_i^j|j=1,2,\dots N\right\}$ and $\left\{\mathbf{k}_i^j|j=1,2,\dots N\right\}$ are enforced to be diverse by a diversity loss
\begin{equation}
  \mathcal{L}_{div}=\sum_{m=1}^{N}\sum_{n=1,n\neq m}^{N}|\mathbf{w}_i^m\otimes\mathbf{w}_i^n|^2 + \sum_{m=1}^{N}\sum_{n=1,n\neq m}^{N}|\mathbf{k}_i^m\otimes\mathbf{k}_i^n|^2.
\end{equation}
The overall training objective on domain $i$ is thus given by
\begin{equation}
  \mathcal{L}_i=\mathcal{L}_{det}+\mathcal{L}_{oim}+\lambda_1\mathcal{L}_{attr}+\lambda_2\mathcal{L}_{div}
  \label{l_cps}
\end{equation}
where $\lambda_1$ and $\lambda_2$ are predefined loss weights.

During inference, we calculate the similarity between an input image $I$ and a learned domain as
\begin{equation}
  s_{I\rightarrow i}= \frac{1}{N}\sum_{j=1}^{N}\textbf{q}_{I}\odot\mathbf{w}_i^j\otimes\mathbf{k}_i^j
\end{equation}
where $\textbf{q}_{I}=F(I)$. Thus the domain index $d$ of the selected prompts $P_{d}^l$ is given by
\begin{equation}
  d=\arg\max_{i}(\left\{s_{I\rightarrow i}|i=1,2,\dots, D\right\})
\end{equation}
where $D$ is the number of learned domains.

We also note that the learning of diverse domain attributes shares a similar technique with CODA-P \cite{coda}. Yet CODA-P is designed for class incremental learning while our work deals with a domain incremental learning problem. CODA-P trains without maximizing the matching scores but implicitly learns attention weights to fuse all prompt components through visual prompt tuning \cite{vpt}. Yet we explicitly optimize the matching scores to guarantee adaptively selecting of learned prompts. The attention mechanism in CODA-P \cite{coda} can also be included in the proposed method to further boost the continual learning performance.

\section{Experiments}

\subsection{Datasets and Evaluation Protocol}

\textbf{CUHK-SYSU} \cite{oim} collected 18,184 images from both movies and real-world street snapshots, presenting 96,143 bounding boxes of pedestrians and 8,432 labeled identities in total. The training set contains 11,206 frames with 5532 identities. The testing set selects 2900 query persons and defines different evaluation protocols with varied gallery sizes.

\textbf{PRW} \cite{prw} collected scene images from 6 cameras deployed at a campus. In total, it presents 11,816 frames containing 43,110 pedestrian bounding boxes with 932 recognizable identities. The training subset contains  5,134 frames with 432 identities and the test set contains 6,112 frames. Different from CUHK-SYSU \cite{oim}, the evaluation protocol of PRW takes the full test set as the gallery.

\textbf{MovieNet-PS} \cite{movienet} selected 160K frames of 3,087 identities from 385 movies. The training set keeps persons of 2,087 identities and a test set is with the left 1,000 identities. For training, it presents 3 different settings that preserve at most 10, 30, and 70 instances per identity, resulting in 20K, 54K, and 100K training images in total. For evaluation, the gallery set is constructed in a way similar to \cite{oim} with varying sizes.

\textbf{Evaluation protocol.} For experiments of CPS, we sequentially train the person search model with the three datasets and then test the model performance on each learned domain without knowing the domain ID. Similar to previous person search research \cite{oim,nae,seqnet,pstr,coat}, we evaluate the person search performance in the format of \textbf{mAP} / \textbf{top-1} in the following experiments.
The \textbf{AP} and \textbf{Recall} scores are used to evaluate the model for person detection.
Moreover, the gallery size is closely related to the degree of retrieval difficulty in person search. To examine the overall model performance, we calculate the weighted average of a person search performance metric as
\begin{equation}
  M_{avg} = \frac{G_{c}}{G} M_{c} + \frac{G_{p}}{G} M_{p} + \frac{G_{m}}{G} M_{m}, G=G_c+G_p+G_m
\end{equation}
where $M_{c}$, $M_{p}$, and $M_{m}$ are the performance metrics on CUHK-SYSU, PRW, and MovieNet-PS, respectively. $G_{c}$, $G_{p}$ and $G_{m}$ are the default gallery size of the respective datasets. We also measure the forgetting of the metrics on learned domains in the same way. Formally, the forgetting measurement is given by the performance decay on domains $i (i<D)$ after learning on domain $D$ compared with that when complete learning on domain $i$. By default, the continual learning sequence is set to CUHK-SYSU \cite{oim}$\rightarrow$PRW \cite{prw}$\rightarrow$MovieNet-PS \cite{movienet}.

\subsection{Implementation Details}

For the compositional person search transformer, we use Swin-S \cite{swin} pre-trained on ImageNet-22K \cite{imagenet} as the pre-trained vision transformer. The output size of RoIAlign \cite{maskrcnn} is set to $14\times 14$ to match the size of image features during pre-training. For the pre-training on person detection data, the model is trained for $36$ epochs with a batch size of $16$. The input image is augmented with random horizontal flip and scaling where the shorter side varies from 480 to 800 during training. We use the AdamW optimizer with an initial learning rate of $0.0001$. The learning rate is linearly warmed up during the first 1,000 iterations and decreased by $10$ at the $20^{th}$ epoch. During the continual learning stage, we employ the Adam optimizer with an initial learning rate of $0.0003$. For CUHK-SYSU \cite{oim} and PRW \cite{prw}, we keep the image augmentation with batch size $8$ and resize the images to $1333\times 800$ for evaluation. For MovieNet-PS \cite{movienet}, we instead randomly resize the shorter image side from 160 to 240. The test image size is set to $720\times 240$ following \cite{movienet}. We train the PoPS model for $28$ epochs on CUHK-SYSU \cite{oim} and PRW \cite{prw}, and $14$ epochs on MoviNet-PS. The loss weights $\lambda_1$ and $\lambda_2$ in Equation \ref{l_cps} are both set to $0.1$. The test gallery sizes of CUHK-SYSU and MovieNet-PS are 100 and 2000, respectively. 

\subsection{Continual Person Search}

To validate the effectiveness of the proposed PoPS method, we conduct continual person search evaluation on \textit{three} types of compared methods: (1) sequentially fine-tuned models including `Prompt + FT-seq' that combines the proposed compositional person search transformer with fixed length prompts, and a representative previous person search model SeqNet \cite{seqnet} with the Swin \cite{swin} backbone used in our method; (2) representative continual learning methods applying to the proposed compositional person search transformer; (3) upper-bound model that performs prompt tuning \cite{vpt} with the compositional person search transformer on the union of all domains. We also test to incorporate the attention mechanism \cite{coda} into PoPS. We directly replace each prompt with a prompt component and bind extra attention vectors and keys \cite{coda} while the weighted summation of prompt components is restricted in the same domain.

\begin{table*}[h]
  \caption{ Continual person search performance comparisons between our proposed PoPS and existing methods. We collect both the person retrieval accuracy and forgetting metrics to make a comprehensive understanding of the effectiveness of PoPS. All results are given as mAP / top-1.}
  \centering
  \begin{tabular}{l|cccc|ccc}
    \toprule
    \multirow{2}*{\textbf{Method}}              & \multicolumn{4}{c|}{\textbf{Accuracy} ($\uparrow$)} & \multicolumn{3}{c}{\textbf{Forgetting} ($\downarrow$)}                                                                                                                                                                                \\
    \cline{2-8}
                                                & \textbf{CUHK-SYSU}                                  & \textbf{PRW}                                           & \textbf{MovieNet-PS}             & \textbf{Average}                    & \textbf{CUHK-SYSU}                & \textbf{PRW}                   & \textbf{Average}               \\
    \midrule
    Prompt + FT-seq                             & 83.2 / 85.3                                         & 20.4 / 76.8                                            & 38.6 / 84.3                      & 25.6 / 79.7                         & 9.9 / 8.8                         & 25.7 / 7.6                     & 25.4 / 7.6                     \\
    SeqNet \cite{seqnet} (Swin \cite{swin})-seq & 75.6 / 77.3                                         & 19.9 / 75.6                                            & 42.1 / 87.2                      & 26.3 / 79.4                         & 17.8 / 16.8                       & 25.9 / 9.1                     & 25.8 / 9.2                     \\
    \midrule
    L2P \cite{l2p}                              & 85.0 / 87.1                                         & 21.7 / 77.0                                            & 36.4 / 82.6                      & 26.4 / 79.4                         & 7.8 / 6.7                         & 24.3 / 7.1                     & 24.0 / 7.1                     \\
    DualPrompt \cite{dualprompt}                & 81.7 / 83.0                                         & 18.1 / 71.5                                            & 36.8 / 83.2                      & 23.7 / 75.4                         & 9.3 / 8.7                         & 23.6 / 7.8                     & 23.4 / 7.8                     \\
    CODA-P \cite{coda}                          & 85.9 / 86.7                                         & 24.7 / 77.5                                            & \textbf{40.2} / \underline{85.7} & 29.7 / 80.6                         & 9.7 / 9.0                         & 26.8 / 9.3                     & 26.5 / 9.3                     \\
    S-Prompt \cite{sprompts}                    & 81.3 / 83.6                                         & 16.7 / 69.8                                            & 32.0 / 79.4                      & 21.5 / 73.2                         & 7.2 / 6.3                         & 6.3 / 2.6                      & 6.3 / 2.7                      \\
    \midrule
    \textbf{PoPS}                               & \underline{86.3} / \underline{87.3}                 & \underline{42.8} / \underline{83.3}                    & 35.4 / 82.3                      & \underline{42.0} / \underline{84.1} & \textbf{5.8} / \textbf{6.0}       & \textbf{0.1} / \textbf{0.2}    & \textbf{0.2} / \textbf{0.3}    \\
    \textbf{PoPS} + Attention \cite{coda}       & \textbf{87.5} / \textbf{88.6}                       & \textbf{49.6} / \textbf{85.8}                          & \underline{39.7} / \textbf{85.9} & \textbf{48.2} / \textbf{86.9}       & \underline{6.8} / \underline{6.4} & \underline{0.2} / \textbf{0.2} & \underline{0.3} / \textbf{0.3} \\
    \midrule
    Prompt + upper-bound                        & 91.6 / 92.8                                         & 46.3 / 84.1                                            & 40.1 / 86.4                      & 45.9 / 85.8                         & -                                 & -                              & -                              \\
    \bottomrule
  \end{tabular}
  \label{tab:cps}
\end{table*}

The experimental performances of the aforementioned methods are presented in Table \ref{tab:cps}. It can be observed that our proposed method obtains the best overall accuracy for CPS. The anti-forgetting performance is also shown to be superior. In addition, incorporating the attention mechanism in CODA-P \cite{coda} further improves the accuracy by introducing more learnable parameters. Compared with previous continual learning methods, sequential prompt tuning of the compositional person search transformer performs only slightly inferior on forgetting learned knowledge, while the sequentially fine-tuned SeqNet \cite{seqnet} is marginally behind. We believe the frozen transformer mainly improves the model stability. We also observe that the domain incremental learning methods S-Prompt \cite{sprompts} and our proposed PoPS perform significantly better on anti-forgetting of the PRW \cite{prw} domain, demonstrating the superior of domain incremental learning techniques for CPS.

\subsection{Analytical Studies}

In this subsection, we conduct ablation experiments to understand the impact of the designed modules in PoPS for CPS. For the convenience of notation, we use \textbf{MVN} as the short for \textbf{MovieNet-PS}.

\textbf{The effectiveness of detection pre-training.} To construct an effectively pre-trained person search transformer without exhaustive pre-training from scratch, we design a compositional person search transformer by expanding a pre-trained Swin with a Simple Feature Pyramid. The added sub-network is then pre-trained by a person detection task. To examine the effectiveness of detection pre-training, we test the pre-trained model on the three person search datasets for person detection and collect the results in Table \ref{tab:detp}. We also test a standard detector Faster R-CNN \cite{fasterrcnn} independently trained on the datasets for comparisons. Although the pre-trained PoPS is evaluated in a cross-domain manner, it can be observed the person detection result is still acceptable compared with the fully supervised Faster R-CNN \cite{fasterrcnn}. This suggests that the detection pre-training is effective for preparing a person search transformer for prompt-based continual learning.

\begin{table}[h]
  \caption{ Person detection performance comparisons between the pre-trained compositional person search transformer and a standard Faster R-CNN \cite{fasterrcnn} detector. }
  \centering
  \begin{tabular}{l|c|c|c}
    \toprule
    \multirow{2}*{\textbf{Method}} & \textbf{CUHK-SYSU}          & \textbf{PRW}                & \textbf{MVN}                \\
    \cline{2-4}
                                   & \textbf{AP}/\textbf{Recall} & \textbf{AP}/\textbf{Recall} & \textbf{AP}/\textbf{Recall} \\
    \midrule
    PoPS                           & 81.8/88.0                   & 90.0/95.4                   & 73.7/83.3                   \\
    Faster R-CNN \cite{fasterrcnn} & 87.0/92.9                   & 93.0/96.1                   & 89.4/96.9                   \\
    \bottomrule
  \end{tabular}
  \label{tab:detp}
\end{table}

\textbf{The effect of attribute projection.} To encourage diverse domain attribute learning for adaptive prompt selection, we introduce the attribute projection embeddings $W_i^l$ to reveal the domain attribute information in an input image before matching it with the learned attribute prototypes. To validate the effect of the attribute projection embeddings, we test PoPS with and without them as in Table \ref{tab:attr}. Without the attribute projection embeddings, we directly maximize the similarity between the image feature $\mathbf{q}_{i}$ and its correlated diverse attribute prototypes $K_i^l$ during training. It can be observed that introducing the attribute projection consistently improves model performance on early domains, suggesting that this mechanism enables more robust prompt selection.

\begin{table}[h]
  \caption{Continual person search performance comparisons between PoPS with and without the attribute projection. }
  \centering
  \begin{tabular}{l|cccc}
    \toprule
    \textbf{Proj.} & \textbf{CUHK-SYSU} & \textbf{PRW} & \textbf{MVN} & \textbf{Average} \\
    \midrule
    $\times$       & 81.3 / 82.3        & 40.9 / 82.0  & 35.5 / 82.1  & 40.6 / 83.0      \\
    $\checkmark$   & 86.3 / 87.3        & 42.8 / 83.3  & 35.4 / 82.3  & 42.0 / 84.1      \\
    \bottomrule
  \end{tabular}
  \label{tab:attr}
\end{table}

\textbf{Comparison between deep prompt and shallow prompt.} As the vision transformers \cite{vit,swin} are formulated in a multi-layer architecture, it is also important to explore which layers to insert the prompts. For this, VPT \cite{vpt} test both shallow (prompts for only the first layer) and deep (prompts for every layer) prompts with ViT \cite{vit}. Different from ViT \cite{vit}, Swin \cite{swin} is composed of 4 stages where each of them contains several layers. We therefore test shallow and deep prompts upon stages for PoPS as in Table \ref{tab:stage}. The results suggest that the model requires deep stage-wise prompts to achieve the best performance. Moreover, adding prompts to the last stage gives a significant improvement in the performance. This is mainly due to that `Stage4' processes the features in a different way, \textit{i.e.} instance-wise, thus requiring extra prompts to adapt well to the person search task.

\begin{table}[h]
  \caption{Continual person search performance of PoPS when prompting different Swin \cite{swin} stages. }
  \centering
  \begin{tabular}{l|cccc}
    \toprule
    \textbf{Stages} & \textbf{CUHK-SYSU} & \textbf{PRW} & \textbf{MVN} & \textbf{Average} \\
    \midrule
    (1,)            & 67.6 / 69.4        & 33.6 / 65.3  & 29.9 / 65.6  & 33.5 / 66.2      \\
    (1,2)           & 69.1 / 70.6        & 34.3 / 66.7  & 30.3 / 66.9  & 34.2 / 67.6      \\
    (1,2,3)         & 72.6 / 73.8        & 36.1 / 70.0  & 31.3 / 69.9  & 35.8 / 70.9      \\
    (1,2,3,4)       & 86.3 / 87.3        & 42.8 / 83.3  & 35.4 / 82.3  & 42.0 / 84.1      \\
    \bottomrule
  \end{tabular}
  \label{tab:stage}
\end{table}

\begin{figure*}[ht]
  \centering
  \begin{subfigure}{0.32\textwidth}
    \centering
    \includegraphics[width=\textwidth]{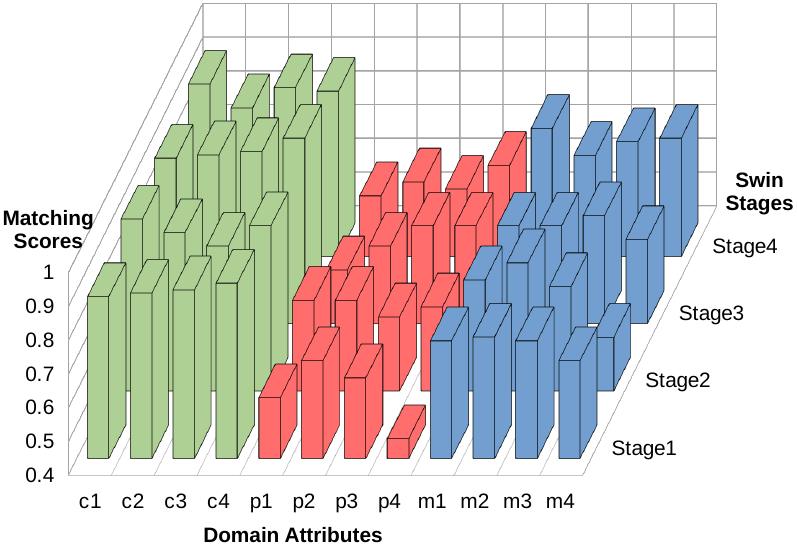}
    \caption{Attribute matching on CUHK-SYSU \cite{oim}.}
    \label{fig:select_c}
  \end{subfigure}
  \begin{subfigure}{0.32\textwidth}
    \centering
    \includegraphics[width=\textwidth]{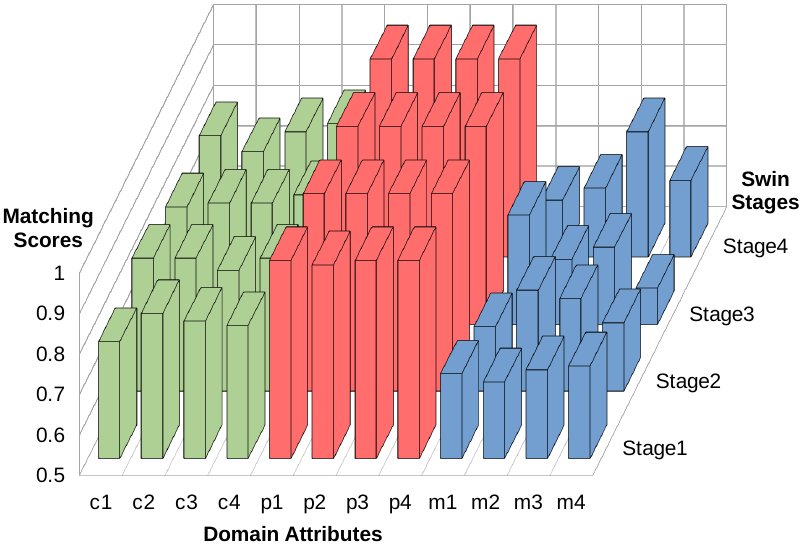}
    \caption{Attribute matching on PRW \cite{prw}.}
    \label{fig:select_p}
  \end{subfigure}
  \begin{subfigure}{0.32\textwidth}
    \centering
    \includegraphics[width=\textwidth]{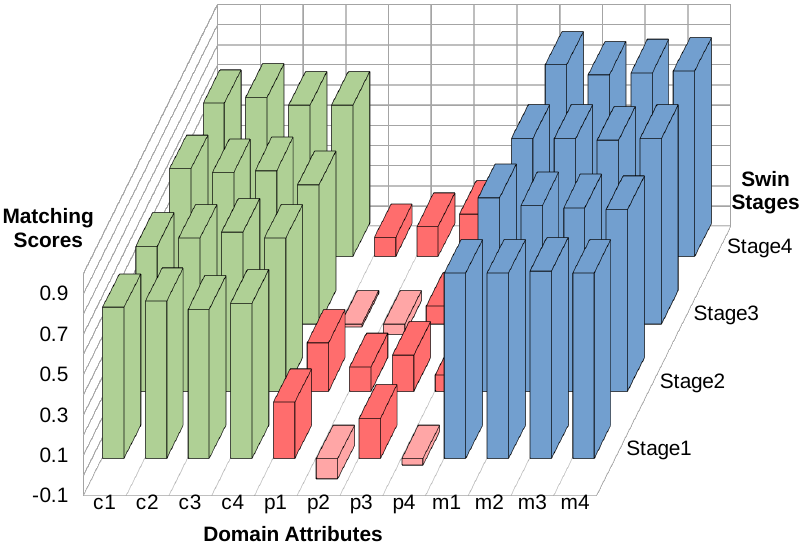}
    \caption{Attribute matching on MVN \cite{movienet}.}
    \label{fig:select_m}
  \end{subfigure}
  \caption{Exampler visualization of prompt selection on CUHK-SYSU \cite{oim}, PRW \cite{prw} and MVN \cite{movienet}. We denote by $\mathrm{c}i$, $\mathrm{p}i$ and $\mathrm{m}i$ the learned $i$-th domain attribute from the three datasets, respectively.}
  \label{fig:select}
\end{figure*}

\textbf{The number of prompts.} With the person search transformer frozen during training, the number of learnable prompts for each transformer layer is also important. To validate the effect of different numbers of learnable prompts, we conduct the experiments in Table \ref{tab:nprompts}. By default, we set the number of prompts $L$ to 16. It can be observed that decreasing the prompts leads to a significant drop in model performance while adding more prompts only slightly improves the results. As the window attention performs self-attention \cite{trans} on relatively short image feature sequences (\textit{e.g.} $7\times 7$), we assume that the inserted prompts should be well balanced with the image feature. Thus the optimal number of prompts should be moderate.

\begin{table}[h]
  \caption{Continual person search performance of PoPS with different numbers of learnable prompts.}
  \centering
  \begin{tabular}{l|cccc}
    \toprule
    $L$ & \textbf{CUHK-SYSU} & \textbf{PRW} & \textbf{MVN} & \textbf{Average} \\
    \midrule
    4      & 81.3 / 82.3        & 39.9 / 82.1  & 33.6 / 80.3  & 39.3 / 82.7      \\
    8      & 84.7 / 86.0        & 40.7 / 82.1  & 34.1 / 80.9  & 30.1 / 82.9      \\
    16     & 86.3 / 87.3        & 42.8 / 83.3  & 35.4 / 82.3  & 42.0 / 84.1      \\
    32     & 86.5 / 87.8        & 43.3 / 83.6  & 36.0 / 82.5  & 42.6 / 84.4      \\
    \bottomrule
  \end{tabular}
  \label{tab:nprompts}
\end{table}

\textbf{The number of learnable domain attributes.} To properly select learned prompts for CPS, we learn to capture $N$ diverse attributes for recognizing each domain. To validate the impact of $N$, we test the CPS performance of PoPS with varied $N$ as in Table \ref{tab:nattr}. It can be observed that learning $4$ diverse domain attributes results in better overall performance. When setting a small $N$, the model may fail to capture sufficient distinct domain features. And the attribute embeddings may overfit on training samples when setting a large $N$ and impedes the person search performance We thus set $N=4$ for each domain by default.

\begin{table}[h]
  \caption{Evaluation of continual person search for PoPS with different numbers of learnable domain attributes. }
  \centering
  \begin{tabular}{l|cccc}
    \toprule
    N & \textbf{CUHK-SYSU} & \textbf{PRW} & \textbf{MVN} & \textbf{Average} \\
    \midrule
    2 & 85.7 / 86.9        & 42.5 / 82.4  & 32.1 / 79.1  & 41.0 / 82.6      \\
    4 & 86.3 / 87.3        & 42.8 / 83.3  & 35.4 / 82.3  & 42.0 / 84.1      \\
    8 & 84.9 / 86.3        & 40.7 / 81.9  & 32.4 / 80.0  & 39.7 / 82.5      \\
    \bottomrule
  \end{tabular}
  \label{tab:nattr}
\end{table}

\textbf{Visualization of prompt selection.} To qualitatively understand the effectiveness of adaptive prompt selection with diverse attribute matching, we visualize the matching scores between input images and learned attributes across different domains. The scores are averaged on $10$ randomly selected test images from the learned domains as in Figure \ref{fig:select}. We also average the scores in different layers within the same stage to obtain stage-level matching scores. It can be observed that the matching scores between the test image and the truly corresponding domain attributes are relatively high compared with the distractors, suggesting the effectiveness of diverse attribute matching. As the learning domains can be highly similar in CPS, we also observe that there always exists a highly similar distracting domain when testing the model.

\section{Conclusion}

This work introduces a new challenging yet practical CPS task that learns from sequentially incoming domains and adaptively completes the person search task for any learned domain. To better balance the stability and plasticity of the model to consistently adapt to new domains without catastrophic forgetting of seen domains, we design a Prompt-based Continual Person Search model (PoPS). To reduce the cost of pre-training a person search transformer from scratch, we first propose a compositional person search transformer that expands a pre-trained hierarchical vision transformer with a Simple Feature Pyramid to enable person localization. We then pre-train the Simple Feature Pyramid with only a moderate number of person detection data to construct a fully pre-trained person search transformer. On top of that, we design a domain incremental prompt pool with a diverse attribute matching module. During training, the prompts are independently learned to encode the domain-oriented knowledge, and a group of paired attribute projections and prototype embeddings are forced to diversely capture distinct domain attributes. This facilitates the adaptive selection of learned prompts by matching an input image with the learned attributes across domains for model inference. For future works, we shall explore designing a more efficient continual learning framework and collecting more realistic domains to better tackle the CPS problem.

\begin{acks}
  This work is supported by the National Natural Science Foundation of China 62276016 and 62372029.
\end{acks}

\bibliographystyle{ACM-Reference-Format}
\bibliography{sample-base}


\begin{thebibliography}{70}


\ifx \showCODEN    \undefined \def \showCODEN     #1{\unskip}     \fi
\ifx \showDOI      \undefined \def \showDOI       #1{#1}\fi
\ifx \showISBNx    \undefined \def \showISBNx     #1{\unskip}     \fi
\ifx \showISBNxiii \undefined \def \showISBNxiii  #1{\unskip}     \fi
\ifx \showISSN     \undefined \def \showISSN      #1{\unskip}     \fi
\ifx \showLCCN     \undefined \def \showLCCN      #1{\unskip}     \fi
\ifx \shownote     \undefined \def \shownote      #1{#1}          \fi
\ifx \showarticletitle \undefined \def \showarticletitle #1{#1}   \fi
\ifx \showURL      \undefined \def \showURL       {\relax}        \fi
\providecommand\bibfield[2]{#2}
\providecommand\bibinfo[2]{#2}
\providecommand\natexlab[1]{#1}
\providecommand\showeprint[2][]{arXiv:#2}

\bibitem[Cao et~al\mbox{.}(2022)]%
        {pstr}
\bibfield{author}{\bibinfo{person}{Jiale Cao}, \bibinfo{person}{Yanwei Pang},
  \bibinfo{person}{Rao~Muhammad Anwer}, \bibinfo{person}{Hisham Cholakkal},
  \bibinfo{person}{Jin Xie}, \bibinfo{person}{Mubarak Shah}, {and}
  \bibinfo{person}{Fahad~Shahbaz Khan}.} \bibinfo{year}{2022}\natexlab{}.
\newblock \showarticletitle{Pstr: End-to-end one-step person search with
  transformers}. In \bibinfo{booktitle}{\emph{Proceedings of the IEEE/CVF
  Conference on Computer Vision and Pattern Recognition}}.
  \bibinfo{pages}{9458--9467}.
\newblock


\bibitem[Caron et~al\mbox{.}(2021)]%
        {dino}
\bibfield{author}{\bibinfo{person}{Mathilde Caron}, \bibinfo{person}{Hugo
  Touvron}, \bibinfo{person}{Ishan Misra}, \bibinfo{person}{Herv{\'e}
  J{\'e}gou}, \bibinfo{person}{Julien Mairal}, \bibinfo{person}{Piotr
  Bojanowski}, {and} \bibinfo{person}{Armand Joulin}.}
  \bibinfo{year}{2021}\natexlab{}.
\newblock \showarticletitle{Emerging properties in self-supervised vision
  transformers}. In \bibinfo{booktitle}{\emph{Proceedings of the IEEE/CVF
  international conference on computer vision}}. \bibinfo{pages}{9650--9660}.
\newblock


\bibitem[Chaudhry et~al\mbox{.}(2018)]%
        {chaudhry2018efficient}
\bibfield{author}{\bibinfo{person}{Arslan Chaudhry},
  \bibinfo{person}{Marc'Aurelio Ranzato}, \bibinfo{person}{Marcus Rohrbach},
  {and} \bibinfo{person}{Mohamed Elhoseiny}.} \bibinfo{year}{2018}\natexlab{}.
\newblock \showarticletitle{Efficient lifelong learning with a-gem}.
\newblock \bibinfo{journal}{\emph{arXiv preprint arXiv:1812.00420}}
  (\bibinfo{year}{2018}).
\newblock


\bibitem[Chen et~al\mbox{.}(2020a)]%
        {hoim}
\bibfield{author}{\bibinfo{person}{Di Chen}, \bibinfo{person}{Shanshan Zhang},
  \bibinfo{person}{Wanli Ouyang}, \bibinfo{person}{Jian Yang}, {and}
  \bibinfo{person}{Bernt Schiele}.} \bibinfo{year}{2020}\natexlab{a}.
\newblock \showarticletitle{Hierarchical Online Instance Matching for Person
  Search}. In \bibinfo{booktitle}{\emph{AAAI}}.
\newblock


\bibitem[Chen et~al\mbox{.}(2020b)]%
        {mgts}
\bibfield{author}{\bibinfo{person}{Di Chen}, \bibinfo{person}{Shanshan Zhang},
  \bibinfo{person}{Wanli Ouyang}, \bibinfo{person}{Jian Yang}, {and}
  \bibinfo{person}{Ying Tai}.} \bibinfo{year}{2020}\natexlab{b}.
\newblock \showarticletitle{Person search by separated modeling and a
  mask-guided two-stream cnn model}.
\newblock \bibinfo{journal}{\emph{IEEE Transactions on Image Processing}}
  \bibinfo{volume}{29} (\bibinfo{year}{2020}), \bibinfo{pages}{4669--4682}.
\newblock


\bibitem[Chen et~al\mbox{.}(2020c)]%
        {nae}
\bibfield{author}{\bibinfo{person}{Di Chen}, \bibinfo{person}{Shanshan Zhang},
  \bibinfo{person}{Jian Yang}, {and} \bibinfo{person}{Bernt Schiele}.}
  \bibinfo{year}{2020}\natexlab{c}.
\newblock \showarticletitle{Norm-aware embedding for efficient person search}.
  In \bibinfo{booktitle}{\emph{Proceedings of the IEEE/CVF conference on
  computer vision and pattern recognition}}. \bibinfo{pages}{12615--12624}.
\newblock


\bibitem[Chen et~al\mbox{.}(2023)]%
        {solider}
\bibfield{author}{\bibinfo{person}{Weihua Chen}, \bibinfo{person}{Xianzhe Xu},
  \bibinfo{person}{Jian Jia}, \bibinfo{person}{Hao Luo},
  \bibinfo{person}{Yaohua Wang}, \bibinfo{person}{Fan Wang},
  \bibinfo{person}{Rong Jin}, {and} \bibinfo{person}{Xiuyu Sun}.}
  \bibinfo{year}{2023}\natexlab{}.
\newblock \showarticletitle{Beyond Appearance: a Semantic Controllable
  Self-Supervised Learning Framework for Human-Centric Visual Tasks}. In
  \bibinfo{booktitle}{\emph{The IEEE/CVF Conference on Computer Vision and
  Pattern Recognition}}.
\newblock


\bibitem[Chen et~al\mbox{.}(2021)]%
        {mocov3}
\bibfield{author}{\bibinfo{person}{Xinlei Chen}, \bibinfo{person}{Saining Xie},
  {and} \bibinfo{person}{Kaiming He}.} \bibinfo{year}{2021}\natexlab{}.
\newblock \showarticletitle{An empirical study of training self-supervised
  vision transformers}. In \bibinfo{booktitle}{\emph{Proceedings of the
  IEEE/CVF international conference on computer vision}}.
  \bibinfo{pages}{9640--9649}.
\newblock


\bibitem[Deng et~al\mbox{.}(2009)]%
        {imagenet}
\bibfield{author}{\bibinfo{person}{Jia Deng}, \bibinfo{person}{Wei Dong},
  \bibinfo{person}{Richard Socher}, \bibinfo{person}{Li-Jia Li},
  \bibinfo{person}{Kai Li}, {and} \bibinfo{person}{Li Fei-Fei}.}
  \bibinfo{year}{2009}\natexlab{}.
\newblock \showarticletitle{Imagenet: A large-scale hierarchical image
  database}. In \bibinfo{booktitle}{\emph{2009 IEEE conference on computer
  vision and pattern recognition}}. Ieee, \bibinfo{pages}{248--255}.
\newblock


\bibitem[Dong et~al\mbox{.}(2020)]%
        {igpn}
\bibfield{author}{\bibinfo{person}{Wenkai Dong}, \bibinfo{person}{Zhaoxiang
  Zhang}, \bibinfo{person}{Chunfeng Song}, {and} \bibinfo{person}{Tieniu Tan}.}
  \bibinfo{year}{2020}\natexlab{}.
\newblock \showarticletitle{Instance guided proposal network for person
  search}. In \bibinfo{booktitle}{\emph{Proceedings of the IEEE/CVF Conference
  on Computer Vision and Pattern Recognition}}. \bibinfo{pages}{2585--2594}.
\newblock


\bibitem[Dosovitskiy et~al\mbox{.}(2020)]%
        {vit}
\bibfield{author}{\bibinfo{person}{Alexey Dosovitskiy}, \bibinfo{person}{Lucas
  Beyer}, \bibinfo{person}{Alexander Kolesnikov}, \bibinfo{person}{Dirk
  Weissenborn}, \bibinfo{person}{Xiaohua Zhai}, \bibinfo{person}{Thomas
  Unterthiner}, \bibinfo{person}{Mostafa Dehghani}, \bibinfo{person}{Matthias
  Minderer}, \bibinfo{person}{Georg Heigold}, \bibinfo{person}{Sylvain Gelly},
  {et~al\mbox{.}}} \bibinfo{year}{2020}\natexlab{}.
\newblock \showarticletitle{An Image is Worth 16x16 Words: Transformers for
  Image Recognition at Scale}. In \bibinfo{booktitle}{\emph{International
  Conference on Learning Representations}}.
\newblock


\bibitem[Fan et~al\mbox{.}(2021)]%
        {mvit}
\bibfield{author}{\bibinfo{person}{Haoqi Fan}, \bibinfo{person}{Bo Xiong},
  \bibinfo{person}{Karttikeya Mangalam}, \bibinfo{person}{Yanghao Li},
  \bibinfo{person}{Zhicheng Yan}, \bibinfo{person}{Jitendra Malik}, {and}
  \bibinfo{person}{Christoph Feichtenhofer}.} \bibinfo{year}{2021}\natexlab{}.
\newblock \showarticletitle{Multiscale vision transformers}. In
  \bibinfo{booktitle}{\emph{Proceedings of the IEEE/CVF international
  conference on computer vision}}. \bibinfo{pages}{6824--6835}.
\newblock


\bibitem[Fu et~al\mbox{.}(2021)]%
        {lup}
\bibfield{author}{\bibinfo{person}{Dengpan Fu}, \bibinfo{person}{Dongdong
  Chen}, \bibinfo{person}{Jianmin Bao}, \bibinfo{person}{Hao Yang},
  \bibinfo{person}{Lu Yuan}, \bibinfo{person}{Lei Zhang},
  \bibinfo{person}{Houqiang Li}, {and} \bibinfo{person}{Dong Chen}.}
  \bibinfo{year}{2021}\natexlab{}.
\newblock \showarticletitle{Unsupervised pre-training for person
  re-identification}. In \bibinfo{booktitle}{\emph{Proceedings of the IEEE/CVF
  conference on computer vision and pattern recognition}}.
  \bibinfo{pages}{14750--14759}.
\newblock


\bibitem[Gao et~al\mbox{.}(2023)]%
        {gao2023unified}
\bibfield{author}{\bibinfo{person}{Qiankun Gao}, \bibinfo{person}{Chen Zhao},
  \bibinfo{person}{Yifan Sun}, \bibinfo{person}{Teng Xi}, \bibinfo{person}{Gang
  Zhang}, \bibinfo{person}{Bernard Ghanem}, {and} \bibinfo{person}{Jian
  Zhang}.} \bibinfo{year}{2023}\natexlab{}.
\newblock \showarticletitle{A unified continual learning framework with general
  parameter-efficient tuning}. In \bibinfo{booktitle}{\emph{Proceedings of the
  IEEE/CVF International Conference on Computer Vision}}.
  \bibinfo{pages}{11483--11493}.
\newblock


\bibitem[Ge et~al\mbox{.}(2022)]%
        {ge2022lifelong}
\bibfield{author}{\bibinfo{person}{Wenhang Ge}, \bibinfo{person}{Junlong Du},
  \bibinfo{person}{Ancong Wu}, \bibinfo{person}{Yuqiao Xian},
  \bibinfo{person}{Ke Yan}, \bibinfo{person}{Feiyue Huang}, {and}
  \bibinfo{person}{Wei-Shi Zheng}.} \bibinfo{year}{2022}\natexlab{}.
\newblock \showarticletitle{Lifelong person re-identification by pseudo task
  knowledge preservation}. In \bibinfo{booktitle}{\emph{Proceedings of the AAAI
  Conference on Artificial Intelligence}}, Vol.~\bibinfo{volume}{36}.
  \bibinfo{pages}{688--696}.
\newblock


\bibitem[Han et~al\mbox{.}(2021a)]%
        {rsiam}
\bibfield{author}{\bibinfo{person}{Chuchu Han}, \bibinfo{person}{Kai Su},
  \bibinfo{person}{Dongdong Yu}, \bibinfo{person}{Zehuan Yuan},
  \bibinfo{person}{Changxin Gao}, \bibinfo{person}{Nong Sang},
  \bibinfo{person}{Yi Yang}, {and} \bibinfo{person}{Changhu Wang}.}
  \bibinfo{year}{2021}\natexlab{a}.
\newblock \showarticletitle{Weakly supervised person search with region siamese
  networks}. In \bibinfo{booktitle}{\emph{Proceedings of the IEEE/CVF
  International Conference on Computer Vision}}. \bibinfo{pages}{12006--12015}.
\newblock


\bibitem[Han et~al\mbox{.}(2021b)]%
        {R-SiamNet}
\bibfield{author}{\bibinfo{person}{Chuchu Han}, \bibinfo{person}{Kai Su},
  \bibinfo{person}{Dongdong Yu}, \bibinfo{person}{Zehuan Yuan},
  \bibinfo{person}{Changxin Gao}, \bibinfo{person}{Nong Sang},
  \bibinfo{person}{Yi Yang}, {and} \bibinfo{person}{Changhu Wang}.}
  \bibinfo{year}{2021}\natexlab{b}.
\newblock \showarticletitle{Weakly Supervised Person Search with Region Siamese
  Networks}. In \bibinfo{booktitle}{\emph{Proceedings of the IEEE/CVF
  International Conference on Computer Vision}}. \bibinfo{pages}{12006--12015}.
\newblock


\bibitem[Han et~al\mbox{.}(2021c)]%
        {dmrnet}
\bibfield{author}{\bibinfo{person}{Chuchu Han}, \bibinfo{person}{Zhedong
  Zheng}, \bibinfo{person}{Changxin Gao}, \bibinfo{person}{Nong Sang}, {and}
  \bibinfo{person}{Yi Yang}.} \bibinfo{year}{2021}\natexlab{c}.
\newblock \showarticletitle{Decoupled and memory-reinforced networks: Towards
  effective feature learning for one-step person search}. In
  \bibinfo{booktitle}{\emph{Proceedings of the AAAI Conference on Artificial
  Intelligence}}. \bibinfo{pages}{1505--1512}.
\newblock


\bibitem[Hayes et~al\mbox{.}(2019)]%
        {hayes2019memory}
\bibfield{author}{\bibinfo{person}{Tyler~L Hayes}, \bibinfo{person}{Nathan~D
  Cahill}, {and} \bibinfo{person}{Christopher Kanan}.}
  \bibinfo{year}{2019}\natexlab{}.
\newblock \showarticletitle{Memory efficient experience replay for streaming
  learning}. In \bibinfo{booktitle}{\emph{2019 International Conference on
  Robotics and Automation (ICRA)}}. IEEE, \bibinfo{pages}{9769--9776}.
\newblock


\bibitem[He et~al\mbox{.}(2017)]%
        {maskrcnn}
\bibfield{author}{\bibinfo{person}{Kaiming He}, \bibinfo{person}{Georgia
  Gkioxari}, \bibinfo{person}{Piotr Doll{\'a}r}, {and} \bibinfo{person}{Ross
  Girshick}.} \bibinfo{year}{2017}\natexlab{}.
\newblock \showarticletitle{Mask r-cnn}. In
  \bibinfo{booktitle}{\emph{Proceedings of the IEEE international conference on
  computer vision}}. \bibinfo{pages}{2961--2969}.
\newblock


\bibitem[Hu et~al\mbox{.}(2018)]%
        {senet}
\bibfield{author}{\bibinfo{person}{Jie Hu}, \bibinfo{person}{Li Shen}, {and}
  \bibinfo{person}{Gang Sun}.} \bibinfo{year}{2018}\natexlab{}.
\newblock \showarticletitle{Squeeze-and-excitation networks}. In
  \bibinfo{booktitle}{\emph{Proceedings of the IEEE conference on computer
  vision and pattern recognition}}. \bibinfo{pages}{7132--7141}.
\newblock


\bibitem[Jia et~al\mbox{.}(2022)]%
        {vpt}
\bibfield{author}{\bibinfo{person}{Menglin Jia}, \bibinfo{person}{Luming Tang},
  \bibinfo{person}{Bor-Chun Chen}, \bibinfo{person}{Claire Cardie},
  \bibinfo{person}{Serge Belongie}, \bibinfo{person}{Bharath Hariharan}, {and}
  \bibinfo{person}{Ser-Nam Lim}.} \bibinfo{year}{2022}\natexlab{}.
\newblock \showarticletitle{Visual prompt tuning}. In
  \bibinfo{booktitle}{\emph{European Conference on Computer Vision}}. Springer,
  \bibinfo{pages}{709--727}.
\newblock


\bibitem[Jung et~al\mbox{.}(2023a)]%
        {jung2023generating}
\bibfield{author}{\bibinfo{person}{Dahuin Jung}, \bibinfo{person}{Dongyoon
  Han}, \bibinfo{person}{Jihwan Bang}, {and} \bibinfo{person}{Hwanjun Song}.}
  \bibinfo{year}{2023}\natexlab{a}.
\newblock \showarticletitle{Generating Instance-level Prompts for
  Rehearsal-free Continual Learning}. In \bibinfo{booktitle}{\emph{Proceedings
  of the IEEE/CVF International Conference on Computer Vision}}.
  \bibinfo{pages}{11847--11857}.
\newblock


\bibitem[Jung et~al\mbox{.}(2023b)]%
        {dap}
\bibfield{author}{\bibinfo{person}{Dahuin Jung}, \bibinfo{person}{Dongyoon
  Han}, \bibinfo{person}{Jihwan Bang}, {and} \bibinfo{person}{Hwanjun Song}.}
  \bibinfo{year}{2023}\natexlab{b}.
\newblock \showarticletitle{Generating Instance-level Prompts for
  Rehearsal-free Continual Learning}. In \bibinfo{booktitle}{\emph{Proceedings
  of the IEEE/CVF International Conference on Computer Vision}}.
  \bibinfo{pages}{11847--11857}.
\newblock


\bibitem[Khan et~al\mbox{.}(2023)]%
        {lgcl}
\bibfield{author}{\bibinfo{person}{Muhammad Gul Zain~Ali Khan},
  \bibinfo{person}{Muhammad~Ferjad Naeem}, \bibinfo{person}{Luc Van~Gool},
  \bibinfo{person}{Didier Stricker}, \bibinfo{person}{Federico Tombari}, {and}
  \bibinfo{person}{Muhammad~Zeshan Afzal}.} \bibinfo{year}{2023}\natexlab{}.
\newblock \showarticletitle{Introducing language guidance in prompt-based
  continual learning}. In \bibinfo{booktitle}{\emph{Proceedings of the IEEE/CVF
  International Conference on Computer Vision}}. \bibinfo{pages}{11463--11473}.
\newblock


\bibitem[Lan et~al\mbox{.}(2018)]%
        {clsa}
\bibfield{author}{\bibinfo{person}{Xu Lan}, \bibinfo{person}{Xiatian Zhu},
  {and} \bibinfo{person}{Shaogang Gong}.} \bibinfo{year}{2018}\natexlab{}.
\newblock \showarticletitle{Person search by multi-scale matching}. In
  \bibinfo{booktitle}{\emph{Proceedings of the European conference on computer
  vision (ECCV)}}. \bibinfo{pages}{536--552}.
\newblock


\bibitem[Li et~al\mbox{.}(2022b)]%
        {li2022domain}
\bibfield{author}{\bibinfo{person}{Junjie Li}, \bibinfo{person}{Yichao Yan},
  \bibinfo{person}{Guanshuo Wang}, \bibinfo{person}{Fufu Yu},
  \bibinfo{person}{Qiong Jia}, {and} \bibinfo{person}{Shouhong Ding}.}
  \bibinfo{year}{2022}\natexlab{b}.
\newblock \showarticletitle{Domain adaptive person search}. In
  \bibinfo{booktitle}{\emph{European Conference on Computer Vision}}. Springer,
  \bibinfo{pages}{302--318}.
\newblock


\bibitem[Li et~al\mbox{.}(2022c)]%
        {uda_ps}
\bibfield{author}{\bibinfo{person}{Junjie Li}, \bibinfo{person}{Yichao Yan},
  \bibinfo{person}{Guanshuo Wang}, \bibinfo{person}{Fufu Yu},
  \bibinfo{person}{Qiong Jia}, {and} \bibinfo{person}{Shouhong Ding}.}
  \bibinfo{year}{2022}\natexlab{c}.
\newblock \showarticletitle{Domain adaptive person search}. In
  \bibinfo{booktitle}{\emph{Proceedings of the European Conference on Computer
  Vision}}. Springer, \bibinfo{pages}{302--318}.
\newblock


\bibitem[Li et~al\mbox{.}(2022a)]%
        {simfpn}
\bibfield{author}{\bibinfo{person}{Yanghao Li}, \bibinfo{person}{Hanzi Mao},
  \bibinfo{person}{Ross Girshick}, {and} \bibinfo{person}{Kaiming He}.}
  \bibinfo{year}{2022}\natexlab{a}.
\newblock \showarticletitle{Exploring plain vision transformer backbones for
  object detection}. In \bibinfo{booktitle}{\emph{European Conference on
  Computer Vision}}. Springer, \bibinfo{pages}{280--296}.
\newblock


\bibitem[Li and Miao(2021)]%
        {seqnet}
\bibfield{author}{\bibinfo{person}{Zhengjia Li} {and} \bibinfo{person}{Duoqian
  Miao}.} \bibinfo{year}{2021}\natexlab{}.
\newblock \showarticletitle{Sequential end-to-end network for efficient person
  search}. In \bibinfo{booktitle}{\emph{Proceedings of the AAAI Conference on
  Artificial Intelligence}}, Vol.~\bibinfo{volume}{35}.
  \bibinfo{pages}{2011--2019}.
\newblock


\bibitem[Lin et~al\mbox{.}(2014)]%
        {coco}
\bibfield{author}{\bibinfo{person}{Tsung-Yi Lin}, \bibinfo{person}{Michael
  Maire}, \bibinfo{person}{Serge Belongie}, \bibinfo{person}{James Hays},
  \bibinfo{person}{Pietro Perona}, \bibinfo{person}{Deva Ramanan},
  \bibinfo{person}{Piotr Doll{\'a}r}, {and} \bibinfo{person}{C~Lawrence
  Zitnick}.} \bibinfo{year}{2014}\natexlab{}.
\newblock \showarticletitle{Microsoft coco: Common objects in context}. In
  \bibinfo{booktitle}{\emph{Computer Vision--ECCV 2014: 13th European
  Conference, Zurich, Switzerland, September 6-12, 2014, Proceedings, Part V
  13}}. Springer, \bibinfo{pages}{740--755}.
\newblock


\bibitem[Liu et~al\mbox{.}(2020)]%
        {dcrnet}
\bibfield{author}{\bibinfo{person}{Jiawei Liu}, \bibinfo{person}{Zheng-Jun
  Zha}, \bibinfo{person}{Richang Hong}, \bibinfo{person}{Meng Wang}, {and}
  \bibinfo{person}{Yongdong Zhang}.} \bibinfo{year}{2020}\natexlab{}.
\newblock \showarticletitle{Dual context-aware refinement network for person
  search}. In \bibinfo{booktitle}{\emph{Proceedings of the 28th ACM
  International Conference on Multimedia}}. \bibinfo{pages}{3450--3459}.
\newblock


\bibitem[Liu et~al\mbox{.}(2021)]%
        {swin}
\bibfield{author}{\bibinfo{person}{Ze Liu}, \bibinfo{person}{Yutong Lin},
  \bibinfo{person}{Yue Cao}, \bibinfo{person}{Han Hu}, \bibinfo{person}{Yixuan
  Wei}, \bibinfo{person}{Zheng Zhang}, \bibinfo{person}{Stephen Lin}, {and}
  \bibinfo{person}{Baining Guo}.} \bibinfo{year}{2021}\natexlab{}.
\newblock \showarticletitle{Swin transformer: Hierarchical vision transformer
  using shifted windows}. In \bibinfo{booktitle}{\emph{Proceedings of the
  IEEE/CVF international conference on computer vision}}.
  \bibinfo{pages}{10012--10022}.
\newblock


\bibitem[Munjal et~al\mbox{.}(2019)]%
        {qeeps}
\bibfield{author}{\bibinfo{person}{Bharti Munjal}, \bibinfo{person}{Sikandar
  Amin}, \bibinfo{person}{Federico Tombari}, {and} \bibinfo{person}{Fabio
  Galasso}.} \bibinfo{year}{2019}\natexlab{}.
\newblock \showarticletitle{Query-guided end-to-end person search}. In
  \bibinfo{booktitle}{\emph{Proceedings of the IEEE/CVF Conference on Computer
  Vision and Pattern Recognition}}. \bibinfo{pages}{811--820}.
\newblock


\bibitem[Oh et~al\mbox{.}(2024)]%
        {oh2024domain}
\bibfield{author}{\bibinfo{person}{Minyoung Oh}, \bibinfo{person}{Duhyun Kim},
  {and} \bibinfo{person}{Jae-Young Sim}.} \bibinfo{year}{2024}\natexlab{}.
\newblock \showarticletitle{Domain Generalizable Person Search Using Unreal
  Dataset}. In \bibinfo{booktitle}{\emph{Proceedings of the AAAI Conference on
  Artificial Intelligence}}, Vol.~\bibinfo{volume}{38}.
  \bibinfo{pages}{4361--4368}.
\newblock


\bibitem[Peng et~al\mbox{.}(2019)]%
        {domainnet}
\bibfield{author}{\bibinfo{person}{Xingchao Peng}, \bibinfo{person}{Qinxun
  Bai}, \bibinfo{person}{Xide Xia}, \bibinfo{person}{Zijun Huang},
  \bibinfo{person}{Kate Saenko}, {and} \bibinfo{person}{Bo Wang}.}
  \bibinfo{year}{2019}\natexlab{}.
\newblock \showarticletitle{Moment matching for multi-source domain
  adaptation}. In \bibinfo{booktitle}{\emph{Proceedings of the IEEE/CVF
  international conference on computer vision}}. \bibinfo{pages}{1406--1415}.
\newblock


\bibitem[Pu et~al\mbox{.}(2021)]%
        {pu2021aka}
\bibfield{author}{\bibinfo{person}{Nan Pu}, \bibinfo{person}{Wei Chen},
  \bibinfo{person}{Yu Liu}, \bibinfo{person}{Erwin~M Bakker}, {and}
  \bibinfo{person}{Michael~S Lew}.} \bibinfo{year}{2021}\natexlab{}.
\newblock \showarticletitle{Lifelong person re-identification via adaptive
  knowledge accumulation}. In \bibinfo{booktitle}{\emph{Proceedings of the
  IEEE/CVF conference on computer vision and pattern recognition}}.
  \bibinfo{pages}{7901--7910}.
\newblock


\bibitem[Pu et~al\mbox{.}(2022)]%
        {pu2022meta}
\bibfield{author}{\bibinfo{person}{Nan Pu}, \bibinfo{person}{Yu Liu},
  \bibinfo{person}{Wei Chen}, \bibinfo{person}{Erwin~M Bakker}, {and}
  \bibinfo{person}{Michael~S Lew}.} \bibinfo{year}{2022}\natexlab{}.
\newblock \showarticletitle{Meta reconciliation normalization for lifelong
  person re-identification}. In \bibinfo{booktitle}{\emph{Proceedings of the
  30th ACM International Conference on Multimedia}}. \bibinfo{pages}{541--549}.
\newblock


\bibitem[Pu et~al\mbox{.}(2023)]%
        {pu2023jaka}
\bibfield{author}{\bibinfo{person}{Nan Pu}, \bibinfo{person}{Zhun Zhong},
  \bibinfo{person}{Nicu Sebe}, {and} \bibinfo{person}{Michael~S Lew}.}
  \bibinfo{year}{2023}\natexlab{}.
\newblock \showarticletitle{A memorizing and generalizing framework for
  lifelong person re-identification}.
\newblock \bibinfo{journal}{\emph{IEEE Transactions on Pattern Analysis and
  Machine Intelligence}} (\bibinfo{year}{2023}).
\newblock


\bibitem[Qin et~al\mbox{.}(2023)]%
        {movienet}
\bibfield{author}{\bibinfo{person}{Jie Qin}, \bibinfo{person}{Peng Zheng},
  \bibinfo{person}{Yichao Yan}, \bibinfo{person}{Rong Quan},
  \bibinfo{person}{Xiaogang Cheng}, {and} \bibinfo{person}{Bingbing Ni}.}
  \bibinfo{year}{2023}\natexlab{}.
\newblock \showarticletitle{MovieNet-PS: a large-scale person search dataset in
  the wild}. In \bibinfo{booktitle}{\emph{ICASSP 2023-2023 IEEE International
  Conference on Acoustics, Speech and Signal Processing (ICASSP)}}. IEEE,
  \bibinfo{pages}{1--5}.
\newblock


\bibitem[Ren et~al\mbox{.}(2015)]%
        {fasterrcnn}
\bibfield{author}{\bibinfo{person}{Shaoqing Ren}, \bibinfo{person}{Kaiming He},
  \bibinfo{person}{Ross Girshick}, {and} \bibinfo{person}{Jian Sun}.}
  \bibinfo{year}{2015}\natexlab{}.
\newblock \showarticletitle{Faster r-cnn: Towards real-time object detection
  with region proposal networks}.
\newblock \bibinfo{journal}{\emph{Advances in neural information processing
  systems}}  \bibinfo{volume}{28} (\bibinfo{year}{2015}).
\newblock


\bibitem[Shao et~al\mbox{.}(2018)]%
        {crowdhuman}
\bibfield{author}{\bibinfo{person}{Shuai Shao}, \bibinfo{person}{Zijian Zhao},
  \bibinfo{person}{Boxun Li}, \bibinfo{person}{Tete Xiao},
  \bibinfo{person}{Gang Yu}, \bibinfo{person}{Xiangyu Zhang}, {and}
  \bibinfo{person}{Jian Sun}.} \bibinfo{year}{2018}\natexlab{}.
\newblock \showarticletitle{Crowdhuman: A benchmark for detecting human in a
  crowd}.
\newblock \bibinfo{journal}{\emph{arXiv preprint arXiv:1805.00123}}
  (\bibinfo{year}{2018}).
\newblock


\bibitem[Shuai et~al\mbox{.}(2022)]%
        {shuai2022id}
\bibfield{author}{\bibinfo{person}{Bing Shuai}, \bibinfo{person}{Xinyu Li},
  \bibinfo{person}{Kaustav Kundu}, {and} \bibinfo{person}{Joseph Tighe}.}
  \bibinfo{year}{2022}\natexlab{}.
\newblock \showarticletitle{Id-free person similarity learning}. In
  \bibinfo{booktitle}{\emph{Proceedings of the IEEE/CVF conference on computer
  vision and pattern recognition}}. \bibinfo{pages}{14689--14699}.
\newblock


\bibitem[Smith et~al\mbox{.}(2023)]%
        {coda}
\bibfield{author}{\bibinfo{person}{James~Seale Smith}, \bibinfo{person}{Leonid
  Karlinsky}, \bibinfo{person}{Vyshnavi Gutta}, \bibinfo{person}{Paola
  Cascante-Bonilla}, \bibinfo{person}{Donghyun Kim}, \bibinfo{person}{Assaf
  Arbelle}, \bibinfo{person}{Rameswar Panda}, \bibinfo{person}{Rogerio Feris},
  {and} \bibinfo{person}{Zsolt Kira}.} \bibinfo{year}{2023}\natexlab{}.
\newblock \showarticletitle{Coda-prompt: Continual decomposed attention-based
  prompting for rehearsal-free continual learning}. In
  \bibinfo{booktitle}{\emph{Proceedings of the IEEE/CVF Conference on Computer
  Vision and Pattern Recognition}}. \bibinfo{pages}{11909--11919}.
\newblock


\bibitem[Sun and Mu(2022)]%
        {sun2022patch}
\bibfield{author}{\bibinfo{person}{Zhicheng Sun} {and} \bibinfo{person}{Yadong
  Mu}.} \bibinfo{year}{2022}\natexlab{}.
\newblock \showarticletitle{Patch-based knowledge distillation for lifelong
  person re-identification}. In \bibinfo{booktitle}{\emph{Proceedings of the
  30th ACM International Conference on Multimedia}}. \bibinfo{pages}{696--707}.
\newblock


\bibitem[Tang et~al\mbox{.}(2023)]%
        {tang2023prompt}
\bibfield{author}{\bibinfo{person}{Yu-Ming Tang}, \bibinfo{person}{Yi-Xing
  Peng}, {and} \bibinfo{person}{Wei-Shi Zheng}.}
  \bibinfo{year}{2023}\natexlab{}.
\newblock \showarticletitle{When prompt-based incremental learning does not
  meet strong pretraining}. In \bibinfo{booktitle}{\emph{Proceedings of the
  IEEE/CVF International Conference on Computer Vision}}.
  \bibinfo{pages}{1706--1716}.
\newblock


\bibitem[Tian et~al\mbox{.}(2022)]%
        {galw}
\bibfield{author}{\bibinfo{person}{Yanling Tian}, \bibinfo{person}{Di Chen},
  \bibinfo{person}{Yunan Liu}, \bibinfo{person}{Shanshan Zhang}, {and}
  \bibinfo{person}{Jian Yang}.} \bibinfo{year}{2022}\natexlab{}.
\newblock \showarticletitle{Grouped Adaptive Loss Weighting for Person Search}.
  In \bibinfo{booktitle}{\emph{Proceedings of the 30th ACM International
  Conference on Multimedia}}. \bibinfo{pages}{6774--6782}.
\newblock


\bibitem[Van~de Ven and Tolias(2019)]%
        {van2019three}
\bibfield{author}{\bibinfo{person}{Gido~M Van~de Ven} {and}
  \bibinfo{person}{Andreas~S Tolias}.} \bibinfo{year}{2019}\natexlab{}.
\newblock \showarticletitle{Three scenarios for continual learning}.
\newblock \bibinfo{journal}{\emph{arXiv preprint arXiv:1904.07734}}
  (\bibinfo{year}{2019}).
\newblock


\bibitem[Vaswani et~al\mbox{.}(2017)]%
        {trans}
\bibfield{author}{\bibinfo{person}{Ashish Vaswani}, \bibinfo{person}{Noam
  Shazeer}, \bibinfo{person}{Niki Parmar}, \bibinfo{person}{Jakob Uszkoreit},
  \bibinfo{person}{Llion Jones}, \bibinfo{person}{Aidan~N Gomez},
  \bibinfo{person}{{\L}ukasz Kaiser}, {and} \bibinfo{person}{Illia
  Polosukhin}.} \bibinfo{year}{2017}\natexlab{}.
\newblock \showarticletitle{Attention is all you need}.
\newblock \bibinfo{journal}{\emph{Advances in neural information processing
  systems}}  \bibinfo{volume}{30} (\bibinfo{year}{2017}).
\newblock


\bibitem[Wang et~al\mbox{.}(2023a)]%
        {wang2023self}
\bibfield{author}{\bibinfo{person}{Benzhi Wang}, \bibinfo{person}{Yang Yang},
  \bibinfo{person}{Jinlin Wu}, \bibinfo{person}{Guo-jun Qi}, {and}
  \bibinfo{person}{Zhen Lei}.} \bibinfo{year}{2023}\natexlab{a}.
\newblock \showarticletitle{Self-similarity driven scale-invariant learning for
  weakly supervised person search}. In \bibinfo{booktitle}{\emph{Proceedings of
  the IEEE/CVF International Conference on Computer Vision}}.
  \bibinfo{pages}{1813--1822}.
\newblock


\bibitem[Wang et~al\mbox{.}(2023b)]%
        {Wang_2023_ICCV}
\bibfield{author}{\bibinfo{person}{Benzhi Wang}, \bibinfo{person}{Yang Yang},
  \bibinfo{person}{Jinlin Wu}, \bibinfo{person}{Guo-jun Qi}, {and}
  \bibinfo{person}{Zhen Lei}.} \bibinfo{year}{2023}\natexlab{b}.
\newblock \showarticletitle{Self-similarity Driven Scale-invariant Learning for
  Weakly Supervised Person Search}. In \bibinfo{booktitle}{\emph{Proceedings of
  the IEEE/CVF International Conference on Computer Vision}}.
  \bibinfo{pages}{1813--1822}.
\newblock


\bibitem[Wang et~al\mbox{.}(2020)]%
        {tcts}
\bibfield{author}{\bibinfo{person}{Cheng Wang}, \bibinfo{person}{Bingpeng Ma},
  \bibinfo{person}{Hong Chang}, \bibinfo{person}{Shiguang Shan}, {and}
  \bibinfo{person}{Xilin Chen}.} \bibinfo{year}{2020}\natexlab{}.
\newblock \showarticletitle{Tcts: A task-consistent two-stage framework for
  person search}. In \bibinfo{booktitle}{\emph{Proceedings of the IEEE/CVF
  Conference on Computer Vision and Pattern Recognition}}.
  \bibinfo{pages}{11952--11961}.
\newblock


\bibitem[Wang et~al\mbox{.}(2024a)]%
        {HiDe}
\bibfield{author}{\bibinfo{person}{Liyuan Wang}, \bibinfo{person}{Jingyi Xie},
  \bibinfo{person}{Xingxing Zhang}, \bibinfo{person}{Mingyi Huang},
  \bibinfo{person}{Hang Su}, {and} \bibinfo{person}{Jun Zhu}.}
  \bibinfo{year}{2024}\natexlab{a}.
\newblock \showarticletitle{Hierarchical decomposition of prompt-based
  continual learning: Rethinking obscured sub-optimality}.
\newblock \bibinfo{journal}{\emph{Advances in Neural Information Processing
  Systems}}  \bibinfo{volume}{36} (\bibinfo{year}{2024}).
\newblock


\bibitem[Wang et~al\mbox{.}(2024b)]%
        {wang2024comprehensive}
\bibfield{author}{\bibinfo{person}{Liyuan Wang}, \bibinfo{person}{Xingxing
  Zhang}, \bibinfo{person}{Hang Su}, {and} \bibinfo{person}{Jun Zhu}.}
  \bibinfo{year}{2024}\natexlab{b}.
\newblock \showarticletitle{A comprehensive survey of continual learning:
  Theory, method and application}.
\newblock \bibinfo{journal}{\emph{IEEE Transactions on Pattern Analysis and
  Machine Intelligence}} (\bibinfo{year}{2024}).
\newblock


\bibitem[Wang et~al\mbox{.}(2022a)]%
        {sprompts}
\bibfield{author}{\bibinfo{person}{Yabin Wang}, \bibinfo{person}{Zhiwu Huang},
  {and} \bibinfo{person}{Xiaopeng Hong}.} \bibinfo{year}{2022}\natexlab{a}.
\newblock \showarticletitle{S-prompts learning with pre-trained transformers:
  An occam’s razor for domain incremental learning}.
\newblock \bibinfo{journal}{\emph{Advances in Neural Information Processing
  Systems}}  \bibinfo{volume}{35} (\bibinfo{year}{2022}),
  \bibinfo{pages}{5682--5695}.
\newblock


\bibitem[Wang et~al\mbox{.}(2022b)]%
        {dualprompt}
\bibfield{author}{\bibinfo{person}{Zifeng Wang}, \bibinfo{person}{Zizhao
  Zhang}, \bibinfo{person}{Sayna Ebrahimi}, \bibinfo{person}{Ruoxi Sun},
  \bibinfo{person}{Han Zhang}, \bibinfo{person}{Chen-Yu Lee},
  \bibinfo{person}{Xiaoqi Ren}, \bibinfo{person}{Guolong Su},
  \bibinfo{person}{Vincent Perot}, \bibinfo{person}{Jennifer Dy},
  {et~al\mbox{.}}} \bibinfo{year}{2022}\natexlab{b}.
\newblock \showarticletitle{Dualprompt: Complementary prompting for
  rehearsal-free continual learning}. In \bibinfo{booktitle}{\emph{European
  Conference on Computer Vision}}. Springer, \bibinfo{pages}{631--648}.
\newblock


\bibitem[Wang et~al\mbox{.}(2022c)]%
        {l2p}
\bibfield{author}{\bibinfo{person}{Zifeng Wang}, \bibinfo{person}{Zizhao
  Zhang}, \bibinfo{person}{Chen-Yu Lee}, \bibinfo{person}{Han Zhang},
  \bibinfo{person}{Ruoxi Sun}, \bibinfo{person}{Xiaoqi Ren},
  \bibinfo{person}{Guolong Su}, \bibinfo{person}{Vincent Perot},
  \bibinfo{person}{Jennifer Dy}, {and} \bibinfo{person}{Tomas Pfister}.}
  \bibinfo{year}{2022}\natexlab{c}.
\newblock \showarticletitle{Learning to prompt for continual learning}. In
  \bibinfo{booktitle}{\emph{Proceedings of the IEEE/CVF Conference on Computer
  Vision and Pattern Recognition}}. \bibinfo{pages}{139--149}.
\newblock


\bibitem[Xiao et~al\mbox{.}(2017)]%
        {oim}
\bibfield{author}{\bibinfo{person}{Tong Xiao}, \bibinfo{person}{Shuang Li},
  \bibinfo{person}{Bochao Wang}, \bibinfo{person}{Liang Lin}, {and}
  \bibinfo{person}{Xiaogang Wang}.} \bibinfo{year}{2017}\natexlab{}.
\newblock \showarticletitle{Joint detection and identification feature learning
  for person search}. In \bibinfo{booktitle}{\emph{Proceedings of the IEEE
  conference on computer vision and pattern recognition}}.
  \bibinfo{pages}{3415--3424}.
\newblock


\bibitem[Xu et~al\mbox{.}(2014)]%
        {PS}
\bibfield{author}{\bibinfo{person}{Yuanlu Xu}, \bibinfo{person}{Bingpeng Ma},
  \bibinfo{person}{Rui Huang}, {and} \bibinfo{person}{Liang Lin}.}
  \bibinfo{year}{2014}\natexlab{}.
\newblock \showarticletitle{Person search in a scene by jointly modeling people
  commonness and person uniqueness}. In \bibinfo{booktitle}{\emph{Proceedings
  of the 22nd ACM International Conference on Multimedia}}.
  \bibinfo{pages}{937--940}.
\newblock


\bibitem[Yan et~al\mbox{.}(2022a)]%
        {cgps}
\bibfield{author}{\bibinfo{person}{Yichao Yan}, \bibinfo{person}{Jinpeng Li},
  \bibinfo{person}{Shengcai Liao}, \bibinfo{person}{Jie Qin},
  \bibinfo{person}{Bingbing Ni}, \bibinfo{person}{Ke Lu}, {and}
  \bibinfo{person}{Xiaokang Yang}.} \bibinfo{year}{2022}\natexlab{a}.
\newblock \showarticletitle{Exploring visual context for weakly supervised
  person search}. In \bibinfo{booktitle}{\emph{Proceedings of the AAAI
  Conference on Artificial Intelligence}}, Vol.~\bibinfo{volume}{36}.
  \bibinfo{pages}{3027--3035}.
\newblock


\bibitem[Yan et~al\mbox{.}(2022b)]%
        {WCGPS}
\bibfield{author}{\bibinfo{person}{Yichao Yan}, \bibinfo{person}{Jinpeng Li},
  \bibinfo{person}{Shengcai Liao}, \bibinfo{person}{Jie Qin},
  \bibinfo{person}{Bingbing Ni}, \bibinfo{person}{Ke Lu}, {and}
  \bibinfo{person}{Xiaokang Yang}.} \bibinfo{year}{2022}\natexlab{b}.
\newblock \showarticletitle{Exploring visual context for weakly supervised
  person search}. In \bibinfo{booktitle}{\emph{Proceedings of the AAAI
  Conference on Artificial Intelligence}}, Vol.~\bibinfo{volume}{36}.
  \bibinfo{pages}{3027--3035}.
\newblock


\bibitem[Yan et~al\mbox{.}(2021)]%
        {alignps}
\bibfield{author}{\bibinfo{person}{Yichao Yan}, \bibinfo{person}{Jinpeng Li},
  \bibinfo{person}{Jie Qin}, \bibinfo{person}{Song Bai},
  \bibinfo{person}{Shengcai Liao}, \bibinfo{person}{Li Liu},
  \bibinfo{person}{Fan Zhu}, {and} \bibinfo{person}{Ling Shao}.}
  \bibinfo{year}{2021}\natexlab{}.
\newblock \showarticletitle{Anchor-free person search}. In
  \bibinfo{booktitle}{\emph{Proceedings of the IEEE/CVF conference on computer
  vision and pattern recognition}}. \bibinfo{pages}{7690--7699}.
\newblock


\bibitem[Yan et~al\mbox{.}(2019)]%
        {ctxg}
\bibfield{author}{\bibinfo{person}{Yichao Yan}, \bibinfo{person}{Qiang Zhang},
  \bibinfo{person}{Bingbing Ni}, \bibinfo{person}{Wendong Zhang},
  \bibinfo{person}{Minghao Xu}, {and} \bibinfo{person}{Xiaokang Yang}.}
  \bibinfo{year}{2019}\natexlab{}.
\newblock \showarticletitle{Learning context graph for person search}. In
  \bibinfo{booktitle}{\emph{Proceedings of the IEEE/CVF conference on computer
  vision and pattern recognition}}. \bibinfo{pages}{2158--2167}.
\newblock


\bibitem[Ye et~al\mbox{.}(2021)]%
        {agw}
\bibfield{author}{\bibinfo{person}{Mang Ye}, \bibinfo{person}{Jianbing Shen},
  \bibinfo{person}{Gaojie Lin}, \bibinfo{person}{Tao Xiang},
  \bibinfo{person}{Ling Shao}, {and} \bibinfo{person}{Steven~CH Hoi}.}
  \bibinfo{year}{2021}\natexlab{}.
\newblock \showarticletitle{Deep learning for person re-identification: A
  survey and outlook}.
\newblock \bibinfo{journal}{\emph{IEEE transactions on pattern analysis and
  machine intelligence}} \bibinfo{volume}{44}, \bibinfo{number}{6}
  (\bibinfo{year}{2021}), \bibinfo{pages}{2872--2893}.
\newblock


\bibitem[Yu et~al\mbox{.}(2023)]%
        {yu2023lifelong}
\bibfield{author}{\bibinfo{person}{Chunlin Yu}, \bibinfo{person}{Ye Shi},
  \bibinfo{person}{Zimo Liu}, \bibinfo{person}{Shenghua Gao}, {and}
  \bibinfo{person}{Jingya Wang}.} \bibinfo{year}{2023}\natexlab{}.
\newblock \showarticletitle{Lifelong person re-identification via knowledge
  refreshing and consolidation}. In \bibinfo{booktitle}{\emph{Proceedings of
  the AAAI Conference on Artificial Intelligence}}, Vol.~\bibinfo{volume}{37}.
  \bibinfo{pages}{3295--3303}.
\newblock


\bibitem[Yu et~al\mbox{.}(2022)]%
        {coat}
\bibfield{author}{\bibinfo{person}{Rui Yu}, \bibinfo{person}{Dawei Du},
  \bibinfo{person}{Rodney LaLonde}, \bibinfo{person}{Daniel Davila},
  \bibinfo{person}{Christopher Funk}, \bibinfo{person}{Anthony Hoogs}, {and}
  \bibinfo{person}{Brian Clipp}.} \bibinfo{year}{2022}\natexlab{}.
\newblock \showarticletitle{Cascade transformers for end-to-end person search}.
  In \bibinfo{booktitle}{\emph{Proceedings of the IEEE/CVF Conference on
  Computer Vision and Pattern Recognition}}. \bibinfo{pages}{7267--7276}.
\newblock


\bibitem[Zhang et~al\mbox{.}(2021b)]%
        {hat}
\bibfield{author}{\bibinfo{person}{Guowen Zhang}, \bibinfo{person}{Pingping
  Zhang}, \bibinfo{person}{Jinqing Qi}, {and} \bibinfo{person}{Huchuan Lu}.}
  \bibinfo{year}{2021}\natexlab{b}.
\newblock \showarticletitle{Hat: Hierarchical aggregation transformers for
  person re-identification}. In \bibinfo{booktitle}{\emph{Proceedings of the
  29th ACM International Conference on Multimedia}}. \bibinfo{pages}{516--525}.
\newblock


\bibitem[Zhang et~al\mbox{.}(2021a)]%
        {kdmot}
\bibfield{author}{\bibinfo{person}{Wei Zhang}, \bibinfo{person}{Lingxiao He},
  \bibinfo{person}{Peng Chen}, \bibinfo{person}{Xingyu Liao},
  \bibinfo{person}{Wu Liu}, \bibinfo{person}{Qi Li}, {and}
  \bibinfo{person}{Zhenan Sun}.} \bibinfo{year}{2021}\natexlab{a}.
\newblock \showarticletitle{Boosting End-to-end Multi-Object Tracking and
  Person Search via Knowledge Distillation}. In
  \bibinfo{booktitle}{\emph{Proceedings of the 29th ACM International
  Conference on Multimedia}}. \bibinfo{pages}{1192--1201}.
\newblock


\bibitem[Zheng et~al\mbox{.}(2017)]%
        {prw}
\bibfield{author}{\bibinfo{person}{Liang Zheng}, \bibinfo{person}{Hengheng
  Zhang}, \bibinfo{person}{Shaoyan Sun}, \bibinfo{person}{Manmohan Chandraker},
  \bibinfo{person}{Yi Yang}, {and} \bibinfo{person}{Qi Tian}.}
  \bibinfo{year}{2017}\natexlab{}.
\newblock \showarticletitle{Person re-identification in the wild}. In
  \bibinfo{booktitle}{\emph{Proceedings of the IEEE conference on computer
  vision and pattern recognition}}. \bibinfo{pages}{1367--1376}.
\newblock


\bibitem[Zhu et~al\mbox{.}(2020)]%
        {ddetr}
\bibfield{author}{\bibinfo{person}{Xizhou Zhu}, \bibinfo{person}{Weijie Su},
  \bibinfo{person}{Lewei Lu}, \bibinfo{person}{Bin Li},
  \bibinfo{person}{Xiaogang Wang}, {and} \bibinfo{person}{Jifeng Dai}.}
  \bibinfo{year}{2020}\natexlab{}.
\newblock \showarticletitle{Deformable DETR: Deformable Transformers for
  End-to-End Object Detection}. In \bibinfo{booktitle}{\emph{International
  Conference on Learning Representations}}.
\newblock


\end{thebibliography}

\clearpage

\section*{Supplementary Experiments}

\begin{table}[h]
	\caption{ Person detection performance comparisons between different pretraining methods. }
	\centering
	\begin{tabular}{l|cc|cc|cc}
		\toprule
		\multirow{2}*{\textbf{Pretrain}} & \multicolumn{2}{c|}{\textbf{CUHK-SYSU}} & \multicolumn{2}{c|}{\textbf{PRW}} & \multicolumn{2}{c}{\textbf{MVN}}                                                   \\
		\cline{2-7}
		& \textbf{AP}                             & \textbf{Recall}                   & \textbf{AP}                      & \textbf{Recall} & \textbf{AP} & \textbf{Recall} \\
		\midrule
		ImageNet-22k \cite{imagenet}     & 81.8                                    & 88.0                              & 90.0                             & 95.4            & 73.7        & 83.3            \\
		ImageNet-1k \cite{imagenet}      & 61.9                                    & 69.0                              & 65.6                             & 73.8            & 59.5        & 80.9            \\
		SOLIDER \cite{solider}           & 82.1                                    & 88.3                              & 89.3                             & 95.5            & 73.9        & 85.3            \\
		\bottomrule
	\end{tabular}
	\label{tab:ptdet}
\end{table}

\begin{table}[h]
	\caption{Continual person search performance comparisons between different pretraining methods. We use IMG-22k as the short for ImageNet-22k for illustration purposes. }
	\centering
	\begin{tabular}{l|cccc}
		\toprule
		\textbf{Pretrain}       & \textbf{CUHK-SYSU} & \textbf{PRW} & \textbf{MVN} & \textbf{Average} \\
		\midrule
		IMG-22k \cite{imagenet} & 86.3 / 87.3        & 42.8 / 83.3  & 35.4 / 82.3  & 42.0 / 84.1      \\
		SOLIDER \cite{solider}  & 89.9 / 90.9        & 29.5 / 76.5  & 21.0 / 60.5  & 28.2 / 72.8      \\
		\bottomrule
	\end{tabular}
	\label{tab:ptps}
\end{table}

\textbf{The effect of vision transformer pre-training.} Following previous peompt-based continual learning methods \cite{l2p,dualprompt,coda,sprompts}, we employ the Swin Transformer \cite{swin} pre-trained on ImageNet-22k \cite{imagenet} to gaurentee the generalization capaility of the pre-trained transformer. To validate the impact of the pre-training, we further test to employ Swin pre-trained with ImageNet-1K \cite{imagenet} in PoPS. We also note that recent work, SOLIDER \cite{solider}, also presents an effective Swin variant pre-trained on large scale unlabeled person images \cite{lup} and achieves superior performances when fine-tuned on downstream tasks. For this, we also test PoPS based on the pre-trained SOLIDER for continual person search.

As is shown in Table \ref{tab:ptdet}, we first compare the pre-training methods on the person detection pre-training stage. It can be observed that although SOLIDER \cite{solider} is trained on large scale person images, the performance on person detection pre-training is similar to the ImageNet-22k pre-trained Swin Transformer. In contrast, the ImageNet-1k pre-trained model falls largely behind, suggesting that the scale of pre-training data is significant to enable effective prompt-based learning. We further conduct continual person search with the SOLIDER \cite{solider} pre-trained model as in Table \ref{tab:ptps}. Although the model with SOLIDER performs more robustly on the CUHK-SYSU dataset, the model fails to fit the more challenging PRW and MovieNet-PS datasets. As the scene images usually contain complex background objects in person search, we hypothesize that the Swin trained with only person images can be less robust than the ImageNet-22k pre-trained version, especially on challenging person search datasets.

\begin{table}[h]
	\caption{ Model performance in low-data environments. }
	\vspace{-3.5mm}
	\centering
	\scalebox{1.0}{
		\begin{tabular}{l|cccc}
			\toprule
			\textbf{Data} & \textbf{CUHK-SYSU} & \textbf{PRW} & \textbf{MVN} & \textbf{Average} \\
			\midrule
			25\%          & 85.7 / 86.5        & 42.0 / 82.1  & 33.1 / 80.3  & 40.4 / 81.7      \\
			50\%          & 86.4 / 87.5        & 41.3 / 82.9  & 34.6 / 82.0  & 40.2 / 82.7      \\
			100\%         & 86.3 / 87.3        & 42.8 / 83.3  & 35.4 / 82.3 & 42.0 / 84.1      \\
			\bottomrule
	\end{tabular}}
	\label{tab:data}
\end{table}

\textbf{Data efficiency of the compositional person search transformer.} To reduce the cost of pretraining a person search transformer, we intriduce a compositional person search transformer by extending a pretrained vision transformer with a detection sub-network. We then only optimize the detection sub-network by the person detection task to form a pretrained person search transformer. The detection pretrain task reduce the cost of collecting data, and the limited learnable parameters can be optimized with less data. To quantitatively understand the data efficiency of the transformer, we test to pretrain the model with reduced data and then conduct the continual learning process. As shown in Table \ref{tab:data}, largely reducing the amount of pretrain data to $50\%$ or even $25\%$ only slightly hinders the continual person search performance, which demonstrates the data efficiency of compositional person search transformer.

\textbf{Comparison between prepend and prefix prompt tuning.} To enable effective learning of visual prompts, DualPrompt \cite{dualprompt} explores conducting prefix prompt tuning instead of directly prepending visual prompts and obtains improved model performance. CODA-P \cite{coda} also follows the prefix prompt tuning mechanism. As we employ a different vision transformer (Swin \cite{swin} vs ViT \cite{vit}) from those works, we conduct comparisons between prepend and prefix prompt tuning in the proposed PoPS. However, as in Table \ref{tab:tuning}, it can be observed that changing the prompt tuning mechanism barely improves the model performance. For simplicity, we thus evaluate all compared methods with the prepend prompt tuning mechanism.

\begin{table}[h]
	\caption{Continual person search performance comparisons between different prompt tuning methods. }
	\centering
	\begin{tabular}{l|cccc}
		\toprule
		\textbf{Pretrain} & \textbf{CUHK-SYSU} & \textbf{PRW} & \textbf{MVN} & \textbf{Average} \\
		\midrule
		prepend           & 86.3 / 87.3        & 42.8 / 83.3  & 35.4 / 82.3  & 42.0 / 84.1      \\
		prefix            & 85.2 / 86.4        & 43.2 / 83.8  & 34.8 / 81.9  & 41.7 / 83.4      \\
		\bottomrule
	\end{tabular}
	\label{tab:tuning}
\end{table}

\begin{table*}[h]
	\caption{ Continual person detection performance of our proposed PoPS. We collect both the person detection accuracy and forgetting metrics to make a comprehensive understanding of the effectiveness of PoPS. All results are given as AP / Recall.}
	\centering
	\begin{tabular}{l|cccc|ccc}
		\toprule
		\multirow{2}*{\textbf{Method}}        & \multicolumn{4}{c|}{\textbf{Accuracy} ($\uparrow$)} & \multicolumn{3}{c}{\textbf{Forgetting} ($\downarrow$)}                                                                                                                                                                             \\
		\cline{2-8}
		& \textbf{CUHK-SYSU}                                  & \textbf{PRW}                                           & \textbf{MovieNet-PS}                & \textbf{Average}                    & \textbf{CUHK-SYSU}             & \textbf{PRW}                & \textbf{Average}               \\
		\midrule
		Pre-trained \textbf{PoPS}             & 81.8 / 88.0                                         & 90.0 / 95.4                                            & 73.7 / 83.3                         & 81.8 / 88.9                         & -                              & -                           & -                              \\
		\midrule
		Prompt + FT-seq                       & 72.5 / 78.7                                         & 88.3 / 92.4                                            & 85.6 / 95.4                         & 82.1 / 88.8                         & 13.5 / 12.5                    & 5.0 / 5.1                   & 9.3 / 8.8                      \\
		\midrule
		L2P \cite{l2p}                        & 76.2 / 82.8                                         & 89.5 / 94.3                                            & \underline{85.4} / \underline{94.9} & 83.9 / 90.7                         & 10.0 / 8.5                     & 3.5 / 3.1                   & 6.8 / 5.8                      \\
		DualPrompt \cite{dualprompt}          & 79.6 / 85.2                                         & 90.3 / 94.8                                            & 85.1 / 94.4                         & 85.0 / 91.5                         & 6.5 / 5.9                      & 2.4 / 2.4                   & 4.4 / 4.1                      \\
		CODA-P \cite{coda}                    & 80.2 / 87.2                                         & 88.7 / 95.7                                            & \textbf{85.6} / \textbf{95.2}       & 84.8 / 92.7                         & 7.0 / 4.8                      & 5.7 / 2.0                   & 6.4 / 3.4                      \\
		S-Prompt \cite{sprompts}              & 83.5 / 87.9                                         & 89.4 / 94.4                                            & 84.8 / 94.1                         & 85.9 / 92.1                         & 3.0 / 4.1                      & 2.1 / 2.4                   & 2.6 / 3.3                      \\
		\midrule
		\textbf{PoPS}                         & \underline{85.0} / \underline{90.4}                 & \underline{92.5} / \underline{97.2}                    & 84.3 / 94.0                         & \underline{87.3} / \underline{93.9} & 1.0 / 0.6                      & \textbf{0.1} / \textbf{0.1} & \underline{0.6} / \textbf{0.4} \\
		\textbf{PoPS} + Attention \cite{coda} & \textbf{85.6} / \textbf{90.6}                       & \textbf{93.6} / \textbf{97.5}                          & 84.5 / 94.6                         & \textbf{87.9} / \textbf{94.2}       & \textbf{0.9} / \underline{0.7} & \textbf{0.1} / \textbf{0.1} & \textbf{0.5} / \textbf{0.4}    \\
		\midrule
		Prompt + upper-bound                  & 83.9 / 89.9                                         & 92.1 / 97.0                                            & 89.3 / 94.8                         & 88.4 / 93.9                         & -                              & -                           & -                           \\
		\bottomrule
	\end{tabular}
	\label{tab:cdet}
\end{table*}

\begin{figure*}[ht]
	\centering
	\begin{subfigure}{0.42\textwidth}
		\centering
		\includegraphics[width=\textwidth]{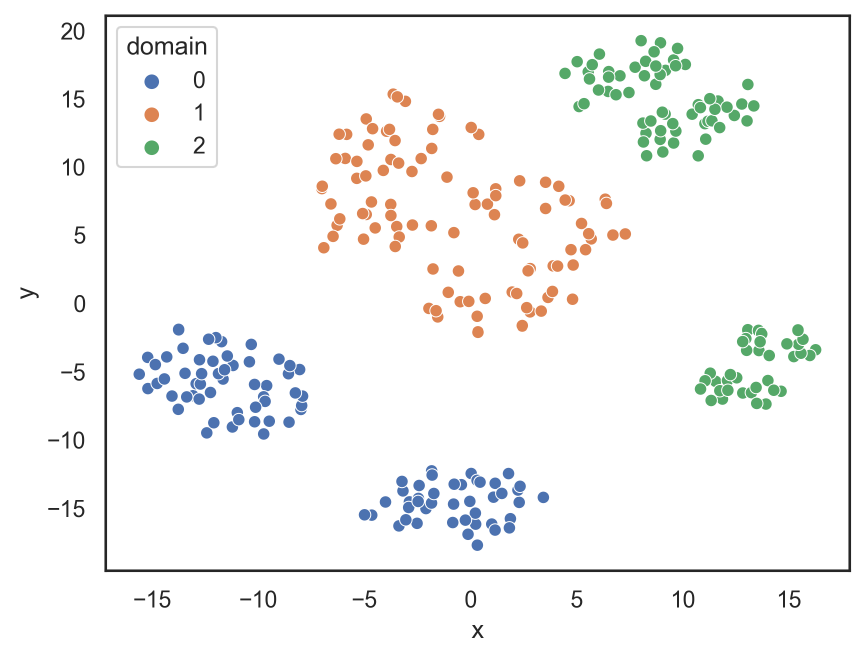}
		\caption{}
		\label{fig:att}
	\end{subfigure}
	\begin{subfigure}{0.42\textwidth}
		\centering
		\includegraphics[width=\textwidth]{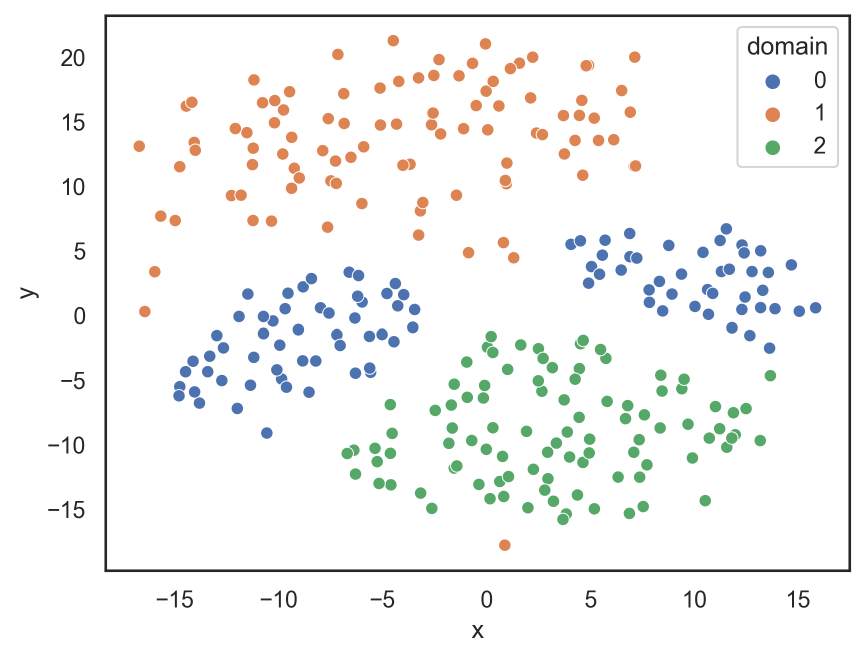}
		\caption{}
		\label{fig:pro}
	\end{subfigure}
	\caption{(a)  T-sne visualization of learned domain attribute prototypes in PoPS. (b) T-sne visualization of learned domain attribute projections in PoPS.}
\end{figure*}

\textbf{Distribution of learned domain attributes.} To qualitatively understand the correlation between learned domain attributes across different person search domains, we conduct t-sne visualization of the learned domain attribute prototypes as well as the attribute projection embeddings as in Figure \ref{fig:att} and Figure \ref{fig:pro}. We refer to CUHK-SYSU \cite{oim}, PRW \cite{prw}, and MovieNet-PS \cite{movienet} as domain 0, 1, and 2, respectively. It can be observed that the learned attribute prototypes effectively capture the distinct differences between learned domains. The attribute projection embeddings also clearly reflect the boundary between different domains, demonstrating the effectiveness of the proposed method.

\textbf{Person detection performance in continual person search.} Person search is a multi-task learning problem that jointly learns to conduct person detection and re-identification \cite{oim}. In addition to the evaluated person search performance, the person detection capability also has an impact to the overall person search accuracy and is affected during the continual learning procedure. We thus evaluate the person detection performance of the proposed method and the compared methods to make a more comprehensive understanding. Different from the evaluation of person retrieval performances, the person detection performances are tested on approximately equal-sized test sets, we thus directly average the results on different domains to obtain an overall performance measurement.

As shown in Table \ref{tab:cdet}, our proposed PoPS consistently achieves superior overall person detection accuracy compared with previous prompt-based continual learning methods \cite{l2p,dualprompt,coda,sprompts}. The anti-forgetting performance is also outstanding on both CUHK-SYSU \cite{oim} and PRW \cite{prw} datasets. It is also worth noting that the overall person detection performance is less hindered by the continual learning procedure compared with the person search performance. This is mainly due to the person detection sub-tasks sharing more common knowledge between different domains while the person retrieval sub-task requires more sophisticated domain-specific knowledge. We also observe that PoPS even performs better than the jointly trained upper-bound. This is mainly caused by the annotation bias in different datasets, \textit{e.g.} some of the small background persons are not annotated in CUHK-SYSU \cite{oim} but annotated in other datasets, which may confuse the model during training.










\end{document}